\documentclass[conference]{IEEEtran}
\usepackage{graphicx}
\usepackage{times}
\usepackage{dsfont}
\usepackage{amsmath,amssymb}
\usepackage[numbers]{natbib}
\usepackage{multicol}
\usepackage{siunitx}
\usepackage[bookmarks=true]{hyperref}
\usepackage{cleveref}
\usepackage{xspace}
\usepackage{xurl}
\usepackage{capt-of}
\usepackage[usenames,dvipsnames]{xcolor}
\usepackage{booktabs}  
\usepackage{threeparttable}

\definecolor{mydarkblue}{rgb}{0,0.08,0.45}
\definecolor{mydarkgreen}{RGB}{0, 139, 69}
\definecolor{mygreen2}{RGB}{0 205 0}
\definecolor{mybrown}{RGB}{139 69 19}
\hypersetup{
	colorlinks=true,
	linkcolor=blue,
	urlcolor=magenta,
	citecolor=mygreen2,
}

\newcommand{\method}{ABS\xspace}

\newcommand{\Revise}[1]{{#1}}

\begin{document}

\title{Agile But Safe: Learning Collision-Free \\High-Speed Legged Locomotion}



\author{\authorblockN{Tairan He\textsuperscript{1\dag}
\quad Chong Zhang\textsuperscript{2\dag} \quad Wenli Xiao\textsuperscript{1} \quad Guanqi He\textsuperscript{1} \quad Changliu Liu\textsuperscript{1} \quad Guanya Shi\textsuperscript{1}}
\authorblockA{
\textsuperscript{1}Carnegie Mellon University \quad \textsuperscript{2}ETH Zürich \quad \textsuperscript{\dag}Equal Contributions \\
Page: \href{https://agile-but-safe.github.io}{\texttt{https://agile-but-safe.github.io}} \quad Code: \href{https://github.com/LeCAR-Lab/ABS}{\texttt{https://github.com/LeCAR-Lab/ABS}}
}
}


%

\makeatletter
\let\@oldmaketitle\@maketitle
    \renewcommand{\@maketitle}{\@oldmaketitle
    \centering
    \includegraphics[width=1.0\textwidth]{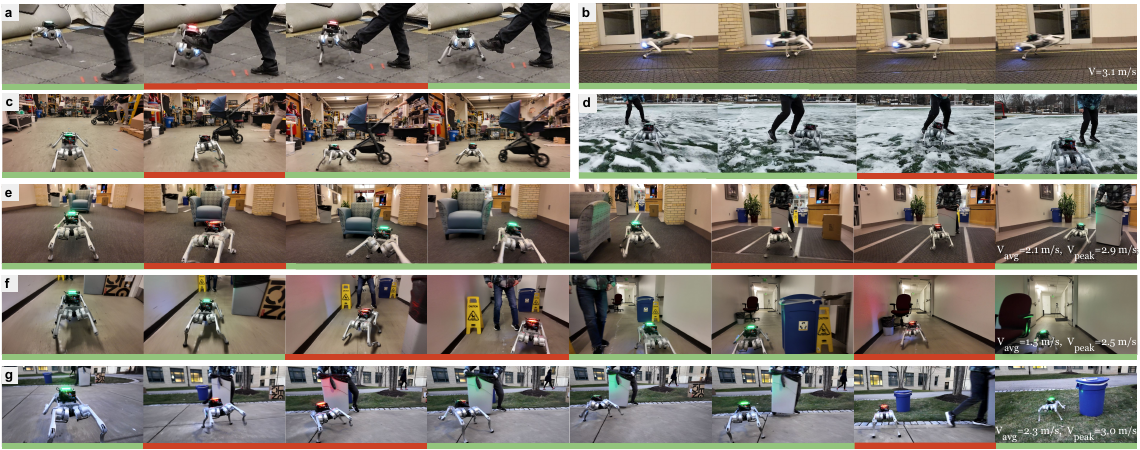}
    \vspace{-0.5cm}
    \captionof{figure}{
    Our proposed framework \textbf{\method} demonstrates agile \emph{and} collision-free locomotion capabilities,  
    where the robot, with fully onboard computation and sensing, can safely navigate through cluttered environments and rapidly react to diverse and dynamic obstacles, both indoors and outdoors.
    \textbf{\method} involves a dual-policy setup: \textcolor{OliveGreen}{\textbf{green}} lines at the bottom indicate the \textcolor{OliveGreen}{\textbf{agile policy}} taking control, and \textcolor{BrickRed}{\textbf{red}} lines indicate the \textcolor{BrickRed}{\textbf{recovery policy}} in operation. The \textcolor{OliveGreen}{\textbf{agile policy}} enables the robot to run fast amidst obstacles, and the \textcolor{BrickRed}{\textbf{recovery policy}} saves the robot from risky cases where the \textcolor{OliveGreen}{\textbf{agile policy}} might fail.\quad
    \textit{\textbf{Subfigures}}: 
    (a) The robot dodges a swinging human leg. 
    (b) The \textcolor{OliveGreen}{\textbf{agile policy}} enables the robot to run at a peak speed of $3.1$~m/s. 
    (c) The robot dodges a moving stroller during high-speed locomotion.
    (d) The robot dodges a moving human in snowy terrain.
    (e) The robot safely navigates in a hall with both static and dynamic obstacles, with an average speed of $2.1$~m/s and a peak speed of $2.9$~m/s.
    (f) The robot avoids obstacles and moving humans in a dim corridor, with an average speed of $1.5$~m/s and a peak speed of $2.5$~m/s. 
    (g) The robot, running outdoors at an average speed of $2.3$~m/s and a peak speed of $3.0$~m/s, avoids both moving and static trash bins and climbs up a grassy slope.
    \textbf{Videos}: see the website. 
    }
    \vspace{-0.2cm} 
    \label{fig:firstpage}
    \setcounter{figure}{1}
  }
\makeatother

\maketitle

\begin{abstract}
Legged robots navigating cluttered environments must be jointly \textit{agile} for efficient task execution and \textit{safe} to avoid collisions with obstacles or humans.
Existing studies either develop conservative controllers ($<1.0$~m/s) to ensure safety, or focus on agility without considering potentially fatal collisions.
This paper introduces Agile But Safe (\method), a learning-based control framework that enables agile and collision-free locomotion for quadrupedal robots. 
\method involves an agile policy to execute agile motor skills amidst obstacles and a recovery policy to prevent failures, collaboratively achieving high-speed and collision-free navigation.
The policy switch in \method is governed by a learned control-theoretic reach-avoid value network, which also guides the recovery policy as an objective function, thereby safeguarding the robot in a closed loop.
The training process involves the learning of the agile policy, the reach-avoid value network, the recovery policy, and an exteroception representation network, all in simulation.
These trained modules can be directly deployed in the real world with onboard sensing and computation, leading to high-speed and collision-free navigation in confined indoor and outdoor spaces with both static and dynamic obstacles (\Cref{fig:firstpage}).
\end{abstract}

\IEEEpeerreviewmaketitle
\vspace{-0.2cm}
\section{Introduction}
\label{sec:introduction}

Agile locomotion of legged robots in cluttered environments presents a non-trivial challenge due to the inherent trade-off between agility and safety, and is crucial for real-world applications that require both robustness and efficiency\Revise{, such as search and rescue~\cite{tranzatto2022cerberus}, disaster response~\cite{sun2020path}, and police robotics~\cite{USAToday2024}}. Existing works typically exhibit limited agility (velocity~$<1$~m/s) to ensure safety~\cite{kim2020vision, dudzik2020robust, buchanan2021perceptive, gaertner2021collision, hoeller2021learning, yang2021learning, chiu2022collision, mattamala2022efficient, liao2023walking, zhang2024resilient}, or focus solely on maximizing agility without considering safety in navigation scenarios~\cite{margolis2022rapid, shin2023actuator}. Our work distinguishes itself by achieving high-speed (max~velocity~$>3$~m/s), collision-free quadrupedal locomotion in cluttered environments.

The agility limitations in existing works in the navigation domain stem from varied factors. Regarding the formulation, some decouple locomotion and navigational planning into two subtasks and build hierarchical systems~\cite{kim2020vision,dudzik2020robust,buchanan2021perceptive,hoeller2021learning,mattamala2022efficient,zhang2024resilient}. Such decoupling not only constrains the controller from the optimal solution~\cite{song2023reaching}, but also results in conservative behaviors to ensure safety, thereby limiting the system from fully unleashing the locomotion agility. This work, instead, learns end-to-end controllers that directly output joint-level actions for collision-free locomotion to reach specified goal positions. 
Our approach is inspired by recent works~\cite{yang2020multi, rudin2022advanced,zhang2023learning,hoeller2023anymal} where robots learn end-to-end controllers to overcome challenging terrains by integrating locomotion with navigation.

Regarding the controller, 
some works employ model-based methods with simplified models, such as model predictive control (MPC) and barrier functions, for guaranteed safety~\cite{gaertner2021collision,chiu2022collision,liao2023walking}. The model mismatch and potential constraint violations such as slippage, together with the online computational burden, limit these controllers from agile motions and stable deployment in the wild~\cite{gaertner2021collision, jenelten2019dynamic, lee2020learning}. On the other hand, recent progress of model-free reinforcement learning (RL) in legged locomotion has demonstrated remarkable agile motor skills that model-based controllers have not achieved~\cite{margolis2022rapid,shin2023actuator,hoeller2023anymal,lee2020learning,miki2022learning,jenelten2023dtc,zhuang2023robot,ma2023learning,yang2023cajun,cheng2023extreme},
although potentially unsafe in cluttered environments. We harness the flexibility and agility of model-free RL and further safeguard it using control-theoretic tools.

Named \method, our framework goes beyond a single RL policy. First, we have a model-free perceptive agile policy that incorporates collision avoidance into locomotion, as presented in \Cref{sec:agilepolicy}, enabling our Go1 robot to achieve peak speeds up to $3.1$~m/s while being aware of collisions. However, the RL policy does not guarantee safety, so we safeguard the robot with another recovery policy (see \Cref{sec:recoverypolicy}) when the agile policy may fail. 
To decide which policy to take control, we use a policy-conditioned reach-avoid (RA) value network to quantify the risk level of the agile policy. This is inspired by \cite{hsu2021safety} where model-free RA values can be efficiently learned based on the Hamilton-Jacobi reachability theory \cite{bansal2017hamilton}. 
The RA value network is trained by a discounted RA Bellman equation, with data collected by the learned agile policy in simulation.
Beyond being a threshold, the differentiable RA value network also provides gradient information to guide the recovery policy, thus closing the loop, which will be further presented in \Cref{sec:reachavoidvalues}.

To get collision avoidance behaviors that can generalize in different scenarios, we use a low-dimensional exteroceptive feature for policy and RA value training: the traveling distances of several rays cast from the robot to obstacles. In order to achieve this, we additionally train an exteroception representation (or ray-prediction) network with simulated data, which maps depth images to ray distances as detailed in \Cref{sec:perception}. By doing so, we achieve robust collision avoidance in high-speed locomotion with onboard sensing and computation.

Briefly, we identify our contributions as follows:
\begin{enumerate}
    \item A perceptive agile policy for obstacle avoidance in high-speed locomotion with novel training methods.
    \item A novel control-theoretic data-driven method for RA value estimation conditioned on the learned agile policy.
    \item A dual-policy setup where an agile policy and a recovery policy collaborates for high-speed collision-free locomotion, and the RA values govern the policy switch and guide the recovery policy.
    \item An exteroception representation network that predicts low-dimensional obstacle information for generalizable collision avoidance capability.
    \item  Validation of \method's superior safety and state-of-the-art agility amidst obstacles indoors and outdoors (\Cref{fig:firstpage}). 
\end{enumerate}

\section{Related Works}
\label{sec:relatedwork}
\subsection{Agile Legged Locomotion}
Model-based methods such as MPC use simplified models and handcrafted gaits to enable dynamic legged locomotion~\cite{bledt2018cheetah,di2018dynamic, ding2019real,grandia2019frequency,jenelten2022tamols,grandia2023perceptive}. 
Despite their impressive performance in simulation and under laboratory conditions, these methods struggle in the wild due to model mismatch and unexpected slippage \cite{jenelten2019dynamic, lee2020learning}. The online computational burden also limits perceptive model-based controllers from agile motions~\cite{chiu2022collision}.

Recently, RL-based controllers have shown promising results for robust locomotion~\cite{yang2020multi, lee2020learning, miki2022learning, gangapurwala2022rloc} and agile motor skills including high-speed running~\cite{hwangbo2019learning, margolis2022rapid, shin2023actuator}, challenging terrain traversal~\cite{zhang2023learning, hoeller2023anymal, jenelten2023dtc, zhuang2023robot, cheng2023extreme}, jumping~\cite{li2023robust, yang2023cajun}, and fall recovery~\cite{yang2020multi, ma2023learning, wang2023guardians, zhang2022accessibility}. However, existing works on agile locomotion mostly study how to achieve fast speeds for racing or skillful motions to overcome challenging terrains. In cluttered environments, these methods necessitate a high-level navigation module for collision avoidance, which is typically conservative and greatly constrains the motion far below the motor limit~\cite{artplanner, zhang2024resilient}. In contrast, this paper studies agile collision avoidance for versatile navigation.

\subsection{Legged Collision Avoidance}
Classical methods tackle collision avoidance in legged robots with collision-free motion planning~\cite{kim2020vision, dudzik2020robust, buchanan2021perceptive} in the configuration space without considering the robot dynamics, leading to slow and statically stable gaits. %
MPC-based methods~\cite{gaertner2021collision, chiu2022collision, teng2021toward, liao2023walking} integrate planning and control by treating distances to obstacles as optimization constraints. However, they suffer from the aforementioned drawbacks of model-based controllers and slow movements (velocity~$<0.5$~m/s).

Learning-based methods are another choice. \Revise{Some existing works~\cite{hoeller2021learning, seo2022prelude, kareer2022vinl, truong2023rethinking, zhang2024resilient, yokoyama2023asc} train RL-based policies} that output twist commands to be tracked by the locomotion controller, while the velocity commands are bounded by $1$~m/s to ensure safety. However, the decoupling of navigation planning and locomotion control makes high-speed locomotion risky, as the high-level planner is unaware of the low-level tracking error. 

\citet{yang2021learning} instead provides an end-to-end RL-based solution that maps depth images and proprioceptive data directly to joint actions, but the robot can only walk forward and the velocity is limited to $\sim0.4$~m/s.
In contrast, our work deploys an end-to-end agile policy for omnidirectional rapid locomotion with collision avoidance, and safeguards the robot with RA values and a recovery policy. To the best of our knowledge, our work is the first to validate collision-free quadruped locomotion with maximum velocity up to $3.1$~m/s. Even in tight space with dynamic adversarial obstacles, our system can still reach a peak velocity of $2.5$~m/s and an average speed of $1.5$~m/s (\Cref{fig:firstpage} (f)).

\subsection{Safe Reinforcement Learning}
There are two main categories of methods to perform safe RL~\cite{ijcai2023p0763}: 1) \textit{end-to-end} methods and 2) \textit{hierarchical} methods. 
Lagrangian-based methods~\cite{OpenAI2019SafeRL, bhatnagar2012online, liang2018accelerated, tessler2018reward} are the most representative \textit{end-to-end} safe RL methods that solve a primal-dual optimization problem to satisfy safety constraint where the Lagrange multipliers can be optimized along with the policy parameters. However, the constraint is often enforced before convergence, hindering exploration and lowering returns~\cite{OpenAI2019SafeRL}.

\textit{Hierarchical} safe RL methods safeguard unsafe RL actions using structures of underlying dynamics~\cite{dalal2018safe,yang2022safe,xiao2023safe} and control-theoretic safety certificates~\cite{cheng2019end,zhao2021model,nakka2020chance}. These methods typically build on the assumptions of available dynamics or safety certificate functions before learning, which heavily limits the scalability to high-dimensional complex systems.
Some recent works learn safety prediction networks (or safety critics) and safety backup policies to safeguard RL when the safety critics indicate the nominal policy is unsafe~\cite{thananjeyan2021recovery,hsu2023sim}. 
Nevertheless, these frameworks lack interplay between safety critics and backup policies, relying on the demanding assumption that the backup policy can restore safety without explicit optimization to satisfy the safety critics.

Our approach aligns with the \textit{hierarchical} methods, yet it stands out with a distinctive strategy. We focus on estimating the reach-avoid values of the agile policy and feed the reach-avoid values' gradient information back into the system to guide the recovery policy within a closed loop. This innovative approach enables a dynamic and adaptive recovery process. Notably, all our modules are trained in simulation using a model-free approach, enhancing the generalizability and scalability of our method.

\subsection{Reach-Avoid Problems and Hamilton-Jacobi Analysis}
Reach-avoid (RA) problems involve navigating a system to reach a target while avoiding certain undesirable states. Hamilton-Jacobi (HJ) reachability analysis~\cite{bansal2017hamilton} solves this problem by analyzing the associated Hamilton-Jacobi partial differential equation, which provides a set of states that the system must stay out of in order to remain safe. 

HJ reachability analysis faces computational challenges, which escalate exponentially with the system's dimensionality~\cite{chen2016fast}. Recent learning-based methods~\cite{bansal2021deepreach} try to scale HJ reachability analysis to high-dimensional systems by learning value networks that satisfy the associated HJ partial differential equations and constraints. However, they still require explicit system Hamiltonian expression before learning. 

Our method builds on another line of works~\cite{fisac2019bridging,hsu2021safety} that leverage contraction properties to derive a time-discounted reach-avoid Bellman equation. However, contrary to previous works that learn policy-agnostic RA values during RL training, we instead learn a policy-conditioned RA value network.
This not only reduces the computational burden by avoiding the identifiability issue of the global RA set but also best suits our trained agile policy. 
\Revise{Similarly, a concurrent work on neural control barrier functions~\cite{so2024train} also applies policy-conditioned safety filters to shield complex systems with reduced complexity.}

\begin{figure*}[t]
    \centering
    \includegraphics[width=\textwidth]{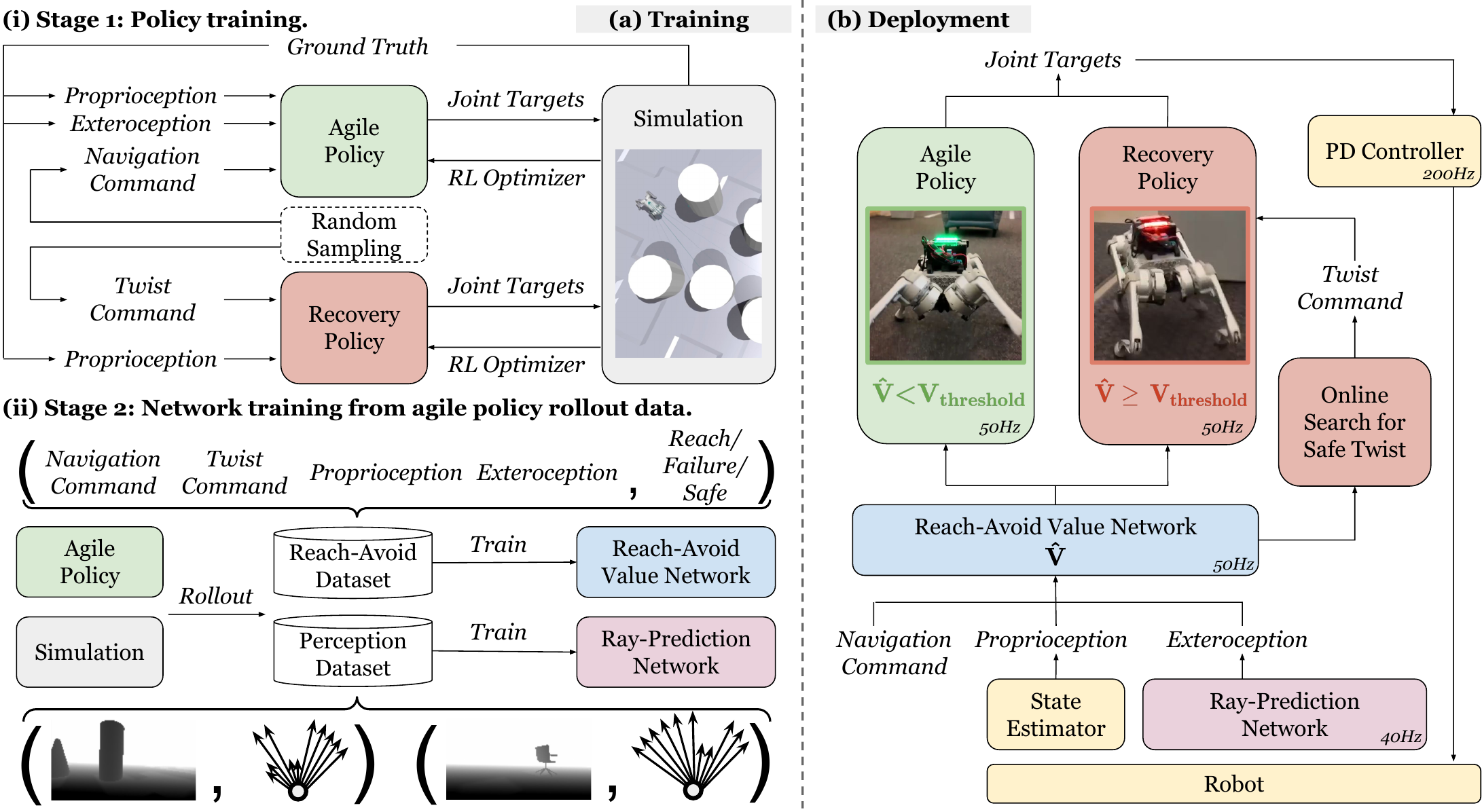}
    \caption{Overview of \method: (a) There are four trained modules within the \method framework: 1) \textcolor{OliveGreen}{\textbf{Agile Policy}} (introduced in \Cref{sec:agilepolicy}) is trained to achieve the maximum agility amidst obstacles; 2) \textcolor{RoyalBlue}{\textbf{Reach-Avoid Value Network}} (introduced in \Cref{sec:reachavoidvalues}) is trained to predict the RA values conditioned on the agile policy as safety indicators; 3) \textcolor{BrickRed}{\textbf{Recovery Policy}} (introduced in \Cref{sec:recoverypolicy}) is trained to track desired twist commands (2D linear velocity $v_x^c, v_y^c$ and yaw angular velocity $\omega_z^c$) that lower the RA values; 4) \textcolor{Purple}{\textbf{Ray-Prediction Network}} (introduced in \Cref{sec:perception}) is trained to predict ray distances as the policies' exteroceptive inputs given depth images. (b) Illustration of the \method deployment architecture. The dual policy setup switches between the \textit{agile policy} and the \textit{recovery policy} based on the estimated $\hat{V}$ from the \textit{RA value network}: 1) if \textcolor{OliveGreen}{$\hat{V} < V_{\text{threshold}}$}, the \textit{agile policy} is activated to navigate amidst obstacles; 2) if \textcolor{BrickRed}{$\hat{V} \geq V_{\text{threshold}}$}, the \textit{recovery policy} is activated to track twist commands that lower the \textit{RA values} via constrained optimization.}
    \label{fig:ABS}
    \vspace{-2mm}
\end{figure*}

\section{Overview and Preliminaries}
\label{sec:overview}

\subsection{Nomenclature}
\label{subsec:nomen}
 We present important symbols and abbreviations that are used across this paper in \Cref{tab:nomen} for reference.

\begin{table}[t]
\centering
\caption{Important Symbols and Abbreviations}
\label{tab:nomen}
\begin{tabular}{cp{6.2cm}}
\hline
Symbol & Meaning \\ \hline
 $t$ & Time step, converted to time in calculation \\
 $s\in \mathcal{S}$ & State \\
 $a\in \mathcal{A}$ & Action \\
 $o\in \mathcal{O}$ & Observation \\
 $T$ & Time horizon or episode length \\
 $\gamma_{\text{RL}}$ & Discount factor for reinforcement learning\\
 $\gamma_{\text{RA}}$ & Discount factor in reach-avoid Bellman equation\\
 $V_\text{threshold}$ & Reach-avoid value threshold equal to $-\epsilon, \epsilon>0$\\
 $\zeta(\cdot)$ & Function indicating failures \\
 $l(\cdot)$ & Function indicating reaching the target \\
 $V_{\text{RA}^*}^{\pi}(\cdot)$ & Ground-truth reach-avoid values conditioned on policy $\pi$ \\
 $V_\text{RA}^{\pi}(\cdot)$ & Ground-truth discounted reach-avoid values conditioned on policy $\pi$ \\
 $\hat{V}(\cdot)$ & Neural network for discounted policy-conditioned reach-avoid value approximation\\
 $\pi^\text{Agile}$ & Agile policy\\
 $\pi^\text{Recovery}$ & Recovery policy \\
 $v$ & Linear velocity in the base frame \\
 $\omega$ & Angular velocity in the base frame \\
 $c_{f}$ & Foot contact statuses\\
 $g$ & Normalized projected gravity in the base frame \\
 $q,\dot{q},\ddot{q}$ & Joint positions, velocities, and accelerations \\
 $\tau$ & Joint torques \\
 $R$ & Logarithm of ray distances \\
 $d_{\text{goal}}$ & Distance from the robot to the goal \\[1mm]
  $G^c$ & Goal command \\
  $v^c$ & Linear velocity command in the base frame \\
 $\omega^c$ & Angular velocity command in the base frame \\
 $tw^c$ & Twist command \\ 
 ${\rm ReLU}(\cdot)$ & Function clipping negative values to zero~\cite{fukushima1969visual} \\ \hline
 Abbreviation & Full Form \\ \hline
 ABS & Agile but safe \\
 RA &  Reach-avoid \\ 
 RL &  Reinforcement learning \\ 
 MPC & Model predictive control \\ 
 MLP & Multilayer perceptron \\
 \hline
\end{tabular}%
\end{table}

\subsection{Problem Formulation}
\label{sec:problemformulation}
\subsubsection{Dynamics}
Let $s_t \in \mathcal{S}\subset \mathbb{R}^{n_s}$ be the state at time step $t$, where $n_s$ is the dimension of the state space $\mathcal{S}$; $a_t \in \mathcal{A}\subset \mathbb{R}^{n_a}$ be the control input at time step $t$, where $n_a$ is the dimension of the action space $\mathcal{A}$. The system dynamics are defined as: 
\begin{equation}\label{eq:dynamics_fn}
\begin{split}
    &s_{t+1} = f(s_t,a_t), \\
\end{split}
\end{equation}
where $f: \mathcal{S} \times \mathcal{A} \rightarrow \mathcal{S}$ is a function that maps the current robot state and control to the next state. For simplicity, this paper considers deterministic dynamics that can be without an analytical form. 
We denote the robot observations from proprioception and/or exteroception as $o_t = h(s_t)$ where $h: \mathcal{S} \rightarrow \mathcal{O}$ is the sensor mapping.
Detailed observation and action space of the agile policy and recovery policy will be introduced in \Cref{sec:agilepolicy} and \Cref{sec:recoverypolicy}. 

\subsubsection{Goal and Policy}
Goal-conditioned reinforcement learning~\cite{liu2022goal} learns to reach goal states $G\in\Gamma$ via a goal-conditioned policy $\pi: \mathcal{O} \times \Gamma \rightarrow \mathcal{A}$. With the reward function $r : \mathcal{S} \times \mathcal{A} \times {\Gamma} \rightarrow \mathbb{R}$ and the discount factor $\gamma_{\text{RL}}$. The policy is learned to maximize the expected cumulative return over the goal distribution $p_G$:
\begin{equation}\label{eq:goal-object}
        J(\pi) = \mathbb{E}_{\substack{a_t \sim \pi(\cdot|o_t, G), G \sim p_G}} \left [\sum_t \gamma_{\text{RL}}^t r(s_t, a_t, G) \right ] ~.
\end{equation}

\subsubsection{Failure Set, Target Set and Reach-Avoid Set}
We denote the \textit{failure set} $\mathcal{F} \subseteq \mathcal{S}$ as unsafe states (e.g., collision) where the robot is not allowed to enter. The \textit{failure set} can be represented by the zero-sublevel set of a Lipschitz-continuous function $\zeta: \mathcal{S} \rightarrow \mathbb{R}$, i.e., $s \in \mathcal{F} \Leftrightarrow \zeta(s) > 0$. 
The \textit{target set} $\Theta \subset \mathcal{S}$ is defined as desired states (i.e., goal states). Similarly, the \textit{target set} can be represented by the zero-sublevel set of a Lipschitz-continuous function $l: \mathcal{S} \rightarrow \mathbb{R}$, i.e., $s \in \Theta \Leftrightarrow l(s) \leq 0$.
We denote $\xi^\pi_{s_t} (\cdot)$ as the future trajectory rollout from state $s_t$ ($\xi^\pi_{s_t} (0)=s_t$) using policy $\pi$ up to $s_T$.
The \textit{reach-avoid set} conditioned on policy $\pi$ is defined as 
\begin{equation}
\begin{split}
\mathcal{RA}^{\pi}(\Theta; \mathcal{F}) :=  \{s_t \in \mathcal{S} \mid & \xi^\pi_{s_t}(T-t) \in \Theta \wedge \\
& \forall t' \in [0, T-t] , \xi^\pi_{s_t}(t') \notin \mathcal{F} \},
\end{split}
\end{equation}
which represents the set of states governed by policy $\pi$ capable of leading the system to $\Theta$ while consistently avoiding $\mathcal{F}$ in all prior timesteps.

\subsubsection{Reach-Avoid Value and Time-Discounted Reach-Avoid Bellman Equation}
\label{subsubsec:ra-bellman}
We define policy-conditioned reach-avoid values as: $V^{\pi}_{\text{RA}^*}(s) \leq 0 \Leftrightarrow s \in \mathcal{RA}^{\pi}(\Theta; \mathcal{F})$. Following the proof (Appendix A in \cite{hsu2021safety}), it can be easily extended that value function $V^{\pi}_{\text{RA}^*}(s)$ satisfies the fixed-point reach-avoid Bellman equation (our policy-conditioned value function is a special case of the general value function):
\begin{equation}
\begin{split}
V^{\pi}_{\text{RA}^*}(s) = \max \Big \{\zeta(s), \min \big \{l(s), V^{\pi}_{\text{RA}^*}\left( f\left(s,\pi(s)\right) \right) \big \} \Big \}~.
\end{split}
\label{eq:RA_bellman_equation}
\end{equation}
However, there is no assurance that \Cref{eq:RA_bellman_equation} will result in a contraction in the space of value functions. To make it accessible to data-driven approximation, we leverage time-discounted reach-avoid Bellman equation~\cite{hsu2021safety} to make a contraction on the discounted policy-conditioned reach-avoid values $V^{\pi}_{\text{RA}}(s)$ defined as 
\begin{equation}
\begin{split}
V^{\pi}_{\text{RA}}(s) = & \gamma_{\text{RA}} \max \Big \{\zeta(s), \min \big \{l(s), V^{\pi}_{\text{RA}}\left( f\left(s,\pi(s)\right) \right) \big \} \Big \} \\ 
& + (1-\gamma_{\text{RA}}) \max \big \{ l(s), \zeta(s) \big \}~.
\end{split}
\label{eq:RA_timediscounted_bellman_equation}
\end{equation}
Following \cite{hsu2021safety}, it can be shown that  $V^{\pi}_{\text{RA}}(s)$ is always an under-approximation of $V^{\pi}_{\text{RA}^*}(s)$ for $\gamma_{\text{RA}} \in [0, 1)$, and $V^{\pi}_{\text{RA}}(s)$ converges to $V^{\pi}_{\text{RA}^*}(s)$ as  $\gamma_{\text{RA}}$ approaches 1.
Note that the under-approximation of $V_{\text{RA}}^\pi(s)$ to $V_{\text{RA}^*}^\pi(s)$ means that $V_{\text{RA}}^\pi(s) \leq 0 \Rightarrow s \in \mathcal{RA}^{\pi}(\Theta; \mathcal{F})$, which enables that shielding methods on thresholds of $V^{\pi}_{\text{RA}}(s)$ could make the system stay in the control-theoretic \textit{reach-avoid set} $\mathcal{RA}^{\pi}(\Theta; \mathcal{F})$.

\subsection{System Structure}

\label{sec:systemstructure}

As shown in \Cref{fig:ABS}, our proposed \method framework involves a dual-policy setup where the agile policy $\pi^{\text{Agile}}$ and the recovery policy $\pi^{\text{Recovery}}$ work together to enable agile and safe locomotion skills. 
The agile policy performs agile motor skills (up to $3.1$~m/s on Unitree Go1) to navigate the robot based on goal commands (target 2D positions and headings) with basic collision-avoidance ability (see also \Cref{sec:agilepolicy}). 
The recovery policy is responsible for safeguarding the agile policy by rapidly tracking twist commands (2D linear velocity $v_x^c, v_y^c$ and yaw rate $\omega_z^c$) that can avoid collisions (see also \Cref{subsec:usingRA} and \Cref{sec:recoverypolicy}). Both policies output joint targets that are tracked by a PD controller.

During deployment, the policy switch is governed by RA values conditioned on the agile policy, estimated using a neural network $\hat{V}$ (see also \Cref{sec:reachavoidvalues}). With a safety threshold $V_{\text{threshold}}=-\epsilon$ where $\epsilon$ is a small positive number, we have: 
\begin{itemize}
    \item If $\hat{V} \geq V_{\text{threshold}}$, we search for a twist command that drives the robot closer to the goal while maintaining safety based on $\hat{V}$ (see also \Cref{eq:ra_rec_solve}). The recovery policy takes control and tracks the searched twist command.
    \item If $\hat{V} < V_{\text{threshold}}$, the agile policy takes control.
\end{itemize}
We expect the system to activate the agile policy in most time, and use the recovery policy as a safeguard in risky situations until it is safe again for the agile policy, \emph{i.e.}, $\hat{V} < V_{\text{threshold}}$.

For collision avoidance, both the agile policy and the RA value networks need exteroceptive inputs. Inspired by~\cite{hoeller2023anymal, duan2023learning, acero2022learning}, we choose to use a low-dimensional exteroception representation: the distances that 11 rays travel from the robot to obstacles, similar to sparse LiDAR readings. We train a network that maps raw depth images to predicted ray distances (see also \Cref{sec:perception}), and the ray distances serve as part of the observations for the agile policy and the RA value network.

To summarize, as shown in \Cref{fig:ABS} (a), \method needs to train four modules all in simulation:
\begin{enumerate}
    \item The \textit{agile policy} (\Cref{sec:agilepolicy}) is trained via RL to reach the goal without collisions. We design goal-reaching rewards to encourage the most agile motor skills.
    \item The \textit{RA value network} (\Cref{sec:reachavoidvalues}) is trained to indicate the safety for the agile policy. We use a data-driven method to train it based on the RA bellman equation (\Cref{eq:RA_timediscounted_bellman_equation}), 
    and collect data in simulation by rolling out the agile policy.
    \item The \textit{recovery policy} (\Cref{sec:recoverypolicy}) is trained to track twist commands rapidly from high-speed movements.
    \item The \textit{ray-prediction network} (\Cref{sec:perception}) is trained to predict ray distance observations from depth images. We collect synthetic depth images and ray distances in simulation by rolling out the agile policy.
\end{enumerate}
All of the four modules are directly deployed in the real world after training.

\section{Learning Agile Policy}
\label{sec:agilepolicy}
As mentioned in \Cref{sec:systemstructure}, we train an agile policy to achieve high agility amidst obstacles. Previous works on learning agile locomotion typically employ the velocity-tracking formulation~\cite{margolis2022rapid, shin2023actuator}, \emph{i.e.}, to track velocity commands on open, flat terrains. However, designing a navigation planner for these velocity-tracking policies in cluttered environments can be non-trivial. To ensure safety, the planner may have to be conservative and unable to fully unleash the locomotion policy's agility.

Instead, we use the goal-reaching formulation to maximize the agility, inspired by~\cite{rudin2022advanced, zhang2023learning}. Specifically, we train the robot to develop sensorimotor skills that enable it to reach specified goals within the episode time without collisions. The agility is also encouraged by a reward term pursuing high velocity in the base frame. By doing so, the robot naturally learns to achieve maximum agility while avoiding collisions.

This section presents the details of our agile policy learning. A detailed comparison between goal-reaching and velocity-tracking formulations for agility will be presented in \Cref{subsubsec:goal_vs_vel}.

\subsection{Observation Space and Action Space}
The observation space of the agile policy consists of the foot contacts $c_{f\in\{1,2,3,4\}}$, the base angular velocities $\omega$, the projected gravity in the base frame $g$, the goal commands $G^c$ (i.e., the relative position and heading of the goal) in the base frame, the time left of the episode $T-t$, the joint positions $q$, the joint velocities $\dot{q}$, the actions $a$ of the previous frame, and the exteroception (i.e., log values of the ray distances) $R$. Here we omit the step-based timestamps (${t-1}$ for the actions and $t$ otherwise) for brevity. We refer to the collection of all these variables as $o^{\text{Agile}}$. 

Among these observations, only $g$ and $G^c$ require the state estimators for respectively orientation and odometry. All other values are available from raw sensor data without cumulative drifts. The IMU-based orientation estimation for $g$ (i.e., roll and pitch) is usually very accurate, and our policy can effectively handle the odometry drift (as we can even suddenly change the goals in the run, see \Cref{subsec:insSteer}). Therefore, our agile policy is robust to inaccurate state estimators which can be problematic for model-based controllers~\cite{bloesch2013state, jenelten2019dynamic, fahmi2021state}.

The action space of the agile policy consists of 12-d joint targets. A PD controller tracks these joint targets $a$ by converting them to joint torques:
\begin{equation}
    \tau = K_p(a-q) -K_d \dot{q}.
\end{equation}
A fully-connected MLP maps the observations $o^{\text{Agile}}$ to the actions $a$.

\subsection{Rewards}

Our reward function is the summation of multiple terms:
\begin{equation}
    r = r_{\text{penalty}} + r_{\text{task}} + r_{\text{regularization}},
\end{equation}
where each term can be further divided into subterms as follows.

\subsubsection{Penalty Rewards} We use a simple penalty design: 
\begin{equation}
    r_{\text{penalty}} = -100\cdot \mathds{1}(\text{undesired collision}),
\end{equation}
where undesired collision refers to collisions on the base, thighs, and calves, and horizontal collisions on the feet.

\subsubsection{Task Rewards} The task rewards are:
\begin{equation}
    \begin{split}
        r_{\text{task}}&= 60\cdot r_{\text{possoft}} + 60\cdot r_{\text{postight}} + 30\cdot r_{\text{heading}}  \\
        &- 10\cdot r_{\text{stand}} + 10\cdot r_{\text{agile}} - 20\cdot r_{\text{stall}},
    \end{split}
\end{equation}
i.e., a soft position tracking term $r_{\text{possoft}}$ to encourage the exploration for goal reaching, a tight position tracking term $r_{\text{postight}}$ to reinforce the robot to stop at the goal, a heading tracking term $r_{\text{heading}}$ to regulate the robot's heading near the goal, a standing term $r_{\text{stand}}$ to encourage a standing posture at the goal, an agile term $r_{\text{agile}}$ to encourage high velocities, and a stall term $r_{\text{stall}}$ to penalize the waiting behaviors. These terms ensure that the robot should reach the goal with appropriate heading and posture as fast as possible while wasting no time.

To be specific, our tracking terms ($r_{\text{possoft}}$, $r_{\text{postight}}$, $r_{\text{heading}}$) are in the same form as shown below, inspired by \cite{zhang2024resilient} where RL-based navigation planners are learned:
\begin{equation}
    r_{\text{track (possoft/postight/heading) }}=\frac{1}{1+\left\|\frac{\text{ error }}{\sigma}\right\|^2}\cdot \frac{\mathds{1}(t>T-T_r)}{T_r},
\end{equation}
where $\sigma$ normalizes the tracking errors, $T$ is the episode length, and $T_r$ is a time threshold. By doing so, the robot only needs to reach the goal before $T-T_r$ to maximize the tracking rewards, free from explicit motion constraints such as target velocities that may limit the agility. For the soft position tracking, we have $\sigma_{\text{soft}}=2$~\unit{\metre} and $T_r=2$~\unit{\second} with the error being the distance to the goal. For the tight position tracking, we have $\sigma_{\text{tight}}=0.5$~\unit{\metre} and $T_r=1$~\unit{\second}. For the heading tracking, we have $\sigma_{\text{heading}}=1$~\unit{rad} and $T_r=2$~\unit{\second} with the error being the relative yaw angle to the goal heading. We further disable $r_{\text{heading}}$ when the distance to the goal is larger than $\sigma_{\text{soft}}$ so that collision avoidance is not affected.

The standing term is defined as 
\begin{equation}
    r_{\text{stand}} = \lVert q-\bar{q} \rVert_1 \cdot \frac{\mathds{1}(t>T-T_{r,\text{stand}})}{T_{r,\text{stand}}}\cdot\mathds{1}(d_{\text{goal}}<\sigma_{\text{tight}}),
\end{equation}
where $\bar{q}$ is the nominal joint positions for standing, $T_{r,\text{stand}}=1$~\unit{\second}, and $d_{\text{goal}}$ is the distance to the goal. 

The agile term is the core term that encourages the agile locomotion. It is defined as 
\begin{equation}
\begin{split}
    r_{\text{agile}} = \max \big\{\rm{ReLU}(\frac{v_x}{v_{\max}}) \cdot \mathds{1}(\text{correct direction}), & \\
    \mathds{1}(d_{\text{goal}}<\sigma_{\text{tight}})&  \big\},
\end{split}
\end{equation}
where $v_x$ is the forward velocity in the robot base frame, $v_{\max}=4.5$~\unit{\meter/\second} is an upper bound of $v_x$ that cannot be reached \Revise{(based on the hardware datasheet)}, and the ``correct direction" means that the angle between the robot heading and the robot-goal line is smaller than 105\unit{\degree}. To maximize this term, the robot has to either run fast or stay at the goal.

The stall term $r_{\text{stall}}$ is $1$ if the robot stays static when $d_{\text{goal}}>\sigma_{\text{soft}}$ and the robot is not in the ``correct direction''. This term penalizes the robot for time waste.

\subsubsection{Regularization Rewards} The regularization rewards are:
\begin{equation}
\begin{aligned}
        r_{\text{regularization}} = & -2\cdot v_z^2 -0.05\cdot(\omega_x^2+\omega_y^2) -20\cdot(g_x^2+g_y^2) \\
        & \hspace{-1.5cm} -0.0005\cdot \lVert \tau \rVert_2^2 - 20\cdot \sum\nolimits_{i=1}^{12} {\rm{ReLU}}\left(\left|\tau_{i}\right|-0.85\cdot\tau_{i,\lim }\right) \\
        &\hspace{-1.5cm} -0.0005\cdot\lVert \dot{q} \rVert_2^2 -20\cdot \sum\nolimits_{i=1}^{12} {\rm{ReLU}} \left(\left|\dot{q}_{i}\right|-0.9\cdot\dot{q}_{i,\lim }\right) \\
        &\hspace{-1.5cm} -20\cdot \sum\nolimits_{i=1}^{12} {\rm{ReLU}} \left(\left|{q}_{i}\right|-0.95\cdot{q}_{i,\lim }\right) \\
        & \hspace{-1.5cm} -2\times10^{-7}\cdot \lVert \ddot{q}\rVert_2^2 -4\times10^{-6}\cdot \lVert \dot{a}\rVert_2^2 -20\cdot\mathds{1}(\text{fly}) ,
\end{aligned}
\end{equation}
where $\tau$ is the joint torques, $\tau_{\lim}$ is the hardware torque limits, $\dot{q}_{\lim}$ is the hardware joint velocity limits, $q_{\lim}$ is the hardware joint position limits, and ``fly'' refers to when the robot has no contact with the ground. We penalize the ``fly'' cases as they make the robot base uncontrollable, threatening the system's safety.

\subsection{Training in Simulation}
\subsubsection{Simulator} We use the GPU-based Isaac Gym simulator~\cite{makoviychuk2021isaac} which supports us to train 1280 environments in parallel with the PPO algorithm~\cite{schulman2017proximal}.
\subsubsection{Terrains}
\begin{figure}[t]
    \centering
    \includegraphics[width=1.0\columnwidth]{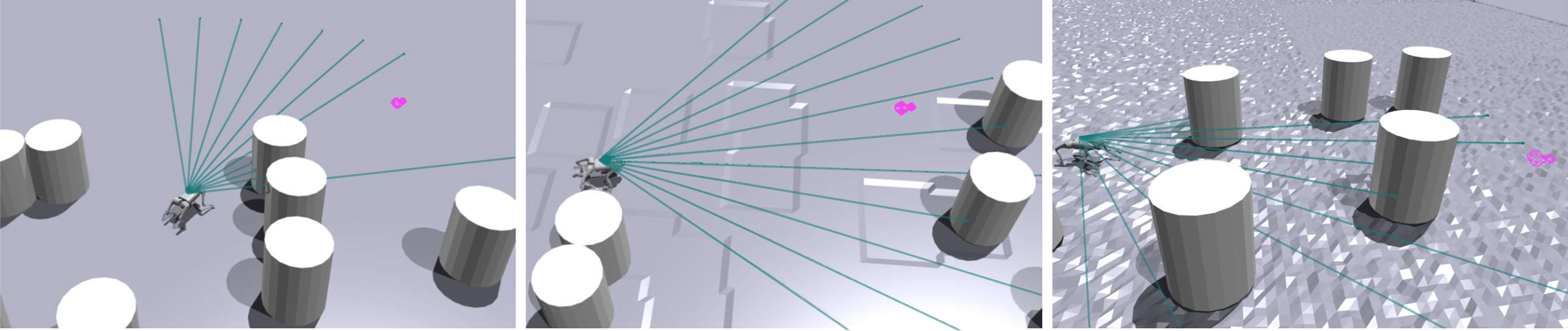}
    \caption{Example training environments. The \textcolor{magenta}{\textbf{magenta}} points indicate the goals, and the \textcolor{BlueGreen}{\textbf{bluegreen}} lines indicate the exteroceptive ray observations. Terrains from left to right: flat, low stumbling blocks, and rough.}
    \label{FIG:terrains}
\end{figure}

We train the agile policy with randomized terrains following a curriculum to facilitate learning. To prevent unstable gaits that over-exploits the simulation dynamics, the terrains are randomly sampled to be flat, rough, or low stumbling blocks, as shown in \Cref{FIG:terrains}. As the difficulty level goes up from 0 to 9, the rough terrains and the stumbling blocks have larger height difference from 0~\unit{cm} to 7~\unit{cm}.

\subsubsection{Obstacles}
We train the policy with cylinders of 40~\unit{cm} radius. For each episode we have 0$\sim$8 obstacles randomly distributed in a 11~\unit{\metre}$\times$ 5~\unit{\meter} rectangle that covers the origin and the goal. To facilitate learning, we also apply a curriculum where higher difficulty levels have more obstacles.

\subsubsection{Domain Randomization}

We do domain randomization~\cite{tobin2017domain} for sim-to-real transfer. The randomized settings are listed in Table~\ref{TAB:domainrand}. Among these few terms, two are critical: the illusion, and the ERFI-50. The illusion makes the policy more robust to unseen geometries such as walls: it overwrites the observed ray distances by random values subject to $\mathcal{U}(d_{\text{goal}}+0.3, \text{ray distance})$ if they are larger than $d_{\text{goal}}+0.3$. The ERFI-50 proposed by~\citet{campanaro2022learning} implicitly models the motor sim-to-real gaps with random torque perturbations, and we add a curriculum in our work to avoid impeding the early stage of learning. We also randomly bias the joint positions to model the motor encoders' offset errors.

\begin{table}[t]
    \centering
\caption{Domain Randomization Settings for Agile Policy Training}
\label{TAB:domainrand}
    \begin{tabular}{ll} 
        \hline
         Term & Value\\ 
        \hline
        \textbf{Observation} & \\
        \quad Illusion & Enabled \\
        \quad Joint position noise & $\mathcal{U}(-0.01,0.01)$~rad\\
        \quad Joint velocity noise & $\mathcal{U}(-1.5,1.5)$~rad/\unit{\second}\\
        \quad Angular velocity noise & $\mathcal{U}(-0.2,0.2)$~rad/\unit{\second}\\
        \quad Projected gravity noise & $ \mathcal{U}(-0.05,0.05)$\\ 
        \quad $\log$(ray distance) noise & $ \mathcal{U}(-0.2,0.2)$\\
        \textbf{Dynamics} & \\
        \quad ERFI-50~\citep{campanaro2022learning} & 0.78~\unit{\newton\metre}$\times$ difficulty level \\
        \quad Friction factor & $ \mathcal{U}(0.4,1.1)$\\ 
        \quad Added base mass & $\mathcal{U}(-1.5,1.5)$~\unit{\kg}\\ 
        \quad Joint position biases & $\mathcal{U}(-0.08,0.08)$~rad\\ 
        \textbf{Episode} &  \\
        \quad Episode length & $\mathcal{U}(7.0,9.0)$~\unit{\second}\\ 
        \quad Initial robot position & $x=0,y=0$ \\ 
        \quad Initial robot yaw & $\mathcal{U}(-\pi,\pi)$~rad \\ 
        \quad Initial robot twist & $\mathcal{U}(-0.5,0.5)$~\unit{\metre/\second} or rad/\unit{\second} \\ 
        \quad Goal Position & $x_{\text{goal}}\sim\mathcal{U}(1.5,7.5)$~\unit{\metre}\\ 
          & $y_{\text{goal}}\sim\mathcal{U}(-2.0,2.0)$~\unit{\metre}\\ 
        \quad Goal Heading & $\arctan2(y_{\text{goal}},x_{\text{goal}})+\mathcal{U}(-0.3,0.3)$~rad\\ 
         \hline
    \end{tabular}
    \vspace{-3mm}
\end{table}

\subsubsection{Curriculum}
As mentioned above, we apply a curriculum where difficulty levels can change the terrains, the obstacle distribution, and the domain randomization. For the assignment of difficulty levels, we follow the design of ~\citet{zhang2023learning}: when an episode terminates, the robot gets promoted to a higher level if $d_{\text{goal}}<\sigma_{\text{tight}}$, and gets demoted to a lower level if $d_{\text{goal}}>\sigma_{\text{soft}}$. If the robot gets promoted at the highest level, it will go to a random level, following~\cite{rudin2022learning}.

\section{Learning and Using Reach-Avoid Values}
\label{sec:reachavoidvalues}
Although the agile policy learns certain collision avoidance behaviors via corresponding rewards, it does not ensure safety. To safeguard the robot, we propose to use RA values to predict the failures, and then a recovery policy can save the robot based on the RA values.

Inspired by \citet{hsu2021safety}, we learn RA values in a model-free way, contrasting typical approaches of model-based reachability analysis~\cite{bansal2021deepreach}. This better suits the model-free RL-based policies. Also different from \cite{hsu2021safety}, we do not learn the global RA values, but make it policy-conditioned, as mentioned in \Cref{subsubsec:ra-bellman}. The learned RA value function will predict only the agile policy's failures based on the observations.

\subsection{Learning RA Values}

To avoid overfitting in high dimensions and make the RA values generalize, we use a reduced set of observations as the inputs of the RA value function:
\begin{equation}
    o^{\text{RA}} = \left[\left[v;\omega \right];G^c_{x,y};R\right],
\end{equation}
\emph{i.e.}, the base twists, the goal $(x,y)$ position in the robot frame, and the exteroception. These components are centroidal observations that significantly influence safety and goal reaching. On the other hand, we don't use joint-level observations (such as $q$ and $\dot{q}$) here because they are high-dimensional and less pertinent to goal reaching.
We train an RA value network $\hat{V}$ to approximate the RA values:
\begin{equation}
    V_{\text{RA}}^{\pi^\text{Agile}}(s)\approx \hat{V}(o^\text{RA}).
\end{equation}

Based on \Cref{eq:RA_timediscounted_bellman_equation}, we minimize the following loss for each episode with gradient descent:
\begin{equation}
    L=\frac{1}{T}\sum_{t=1}^{T} \left(\hat{V}(o^{\text{RA}}_t)-\hat{V}^{\text{target}}\right)^2,
\end{equation}
where 
\begin{equation}
\begin{split}
\hat{V}^{\text{target}} = &
    \gamma_{\text{RA}}\max \Big \{\zeta(s_t), \min \big \{l(s_t), \hat{V}^{\text{old}}(o^{\text{RA}}_{t+1}) \big \} \Big \} \\
    & + (1-\gamma_{\text{RA}}) \max \big \{ l(s_t), \zeta(s_t) \big \},
\end{split}
\end{equation}
and we set the discount factor $\gamma_{\text{RA}}=0.999999$ to best approximate $\mathcal{RA}^{\pi}(\Theta; \mathcal{F})$ since $V^{\pi}_{\text{RA}}(s)$ converges to $V^{\pi}_{\text{RA}^*}(s)$ as  $\gamma_{\text{RA}}$ approaches 1. $\hat{V}^{\text{old}}$ refers to $\hat{V}$ from previous iteration, and we set $\hat{V}^{\text{old}}(o^{\text{RA}}_{T+1})=+\infty$.

Differing from \cite{hsu2021safety}, our approach learns policy-conditioned reach-avoid values instead of solving policy-agnostic global 
reach-avoid value of the entire system dynamics which involves another value minimization problem over $\mathcal{A}$. Our method offers several advantages: 1) simplicity: as highlighted in \Cref{eq:RA_timediscounted_bellman_equation}, this simplicity arises from avoiding the need to solve for the lowest value of the next state across the entire action space. 2) two-stage offline learning: our approach can be learned in a two-stage offline manner. This involves first collecting policy trajectories and then training the policy-conditioned reach-avoid value. This two-stage process enhances stability compared to the online training method presented in \cite{hsu2021safety}.

\subsection{Implementation}
According to~\cite{hsu2021safety}, $l(s)$ and $\zeta(s)$ should be Lipschitz-continuous for theoretical guarantees. In our implementation, we define the $l(s)$ as
\begin{equation}
    l(s) = \tanh{\log \frac{d_{\text{goal}}}{\sigma_{\text{tight}}}},
\end{equation}
thereby making it Lipschitz-continuous, bounding it with $(-1,1)$, and setting $d_{\text{goal}}\le \sigma_{\text{tight}}$ as ``reach".

Regarding failures, we naturally have 
\begin{equation}
\label{eq:failurefunc}
\zeta(s)=2*\mathds{1}(\text{undesired collision})-1.
\end{equation}
However, this definition violates the Lipschitz continuity. Hence, we soften the function in a hindsight way: when an undesired collision happens, the $\zeta$ values for the last 10 timesteps are relabelled to be $-0.8, -0.6, \dots, 0.8, 1.0$.

For RA dataset sampling, we make the obstacles distribute as in the highest difficulty level during the training of the agile policy. We roll out our trained agile policy for 200k episodes, and collect these trajectories for RA learning.

\begin{figure}[t]
    \centering
    \includegraphics[width=1.0\columnwidth]{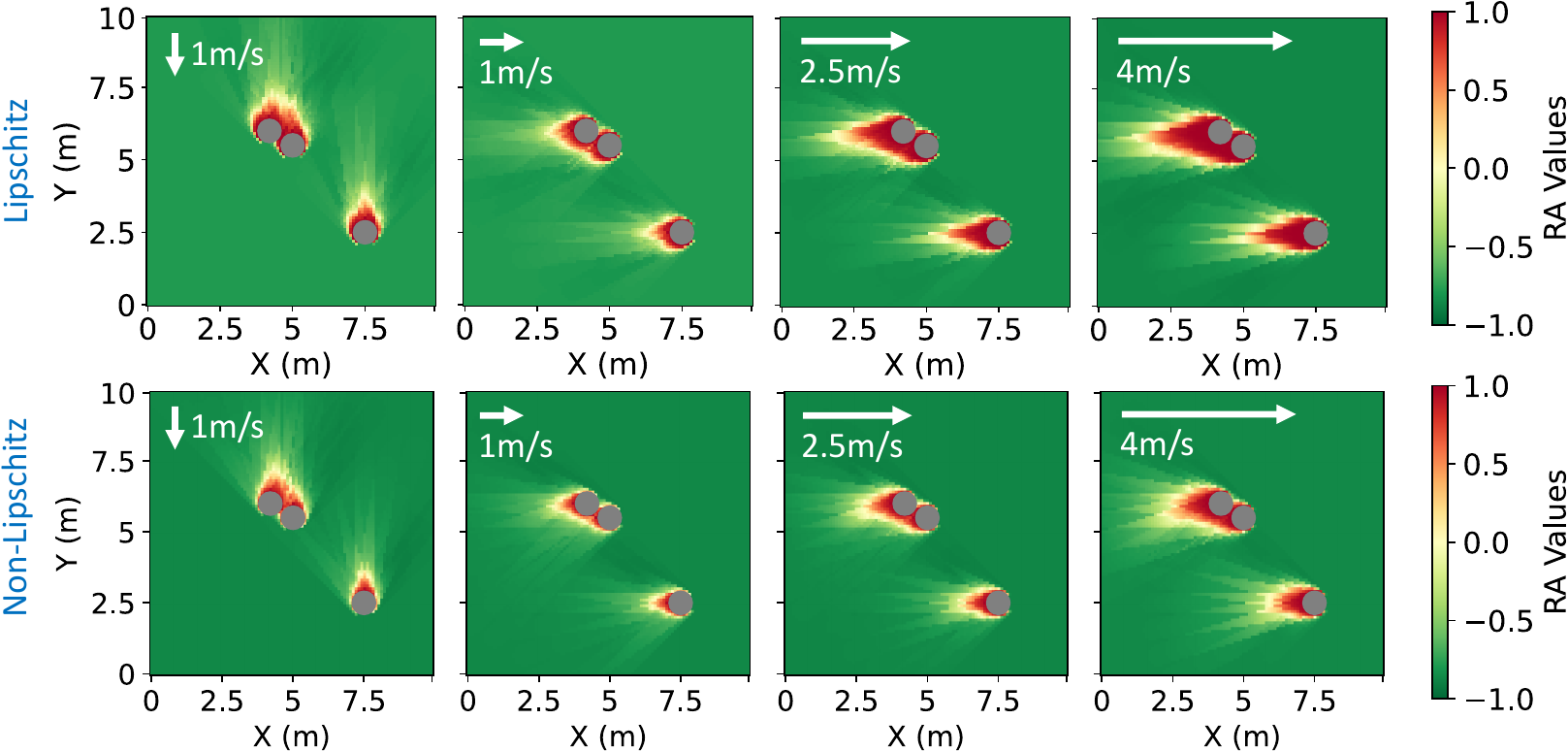}
    \caption{Visualization of $\hat{V}$ with different linear velocities and 2D positions relative to the $3$ fixed obstacles. The angular velocities are set to zero, and the relative goal commands are set to $5$~m ahead of the robot. The grey circles represent the obstacles, and the colors represent the values of $\hat{V}$ at corresponding 2D positions. \Revise{The first row presents the RA values trained with the softened failure function $\zeta$, while the second row uses the raw one in \Cref{eq:failurefunc}. Without softening $\zeta$ to approach the Lipschitz continuity, the value estimation fails to indicate collisions on the sides of obstacles and has local minima in front of the obstacles, compromising safety.}}
    \label{FIG:RA-POS_visualization}
    \vspace{-2mm}
\end{figure}

\Cref{FIG:RA-POS_visualization} visualizes the learned RA values for a specific set of obstacles. As the robot's velocity changes, the landscape of RA values changes accordingly. The sign of the RA values reasonably indicates the safety for the agile policy.


\subsection{Using RA Values for Recovery}
\label{subsec:usingRA}
RA values provide a failure prediction conditioned on the agile policy, and we propose to use RA values to guide the recovery policy. To be specific, the robot decides the optimal twist to avoid collisions using the RA value function, and employs the recovery policy to track these twist commands.
The recovery policy is triggered as a back-up shielding policy if and only if $\hat{V}(o^{\text{RA}})\ge V_\text{threshold}$. We set $V_\text{threshold}=-0.05$ to compensate for learning errors without causing over-conservative shielding.

During recovery, we assume that the recovery policy is well-trained so that the robot twist is close to the command 
\begin{equation}
    tw^c=[v_x^c,v_y^c,0,0,0,\omega_z^c],
\end{equation}
and the robot should try to get closer to the goal if its twist is safe given the goal and the exteroception. Therefore, the twist command is obtained from the optimization:
\begin{equation}
    tw^c = \arg\min d_{\text{goal}}^{\text{future}} \text{ s.t. } \hat{V}([tw^c; G^c_{x,y}; R]) < V_\text{threshold},
\label{eq:ra_rec_solve}
\end{equation}
and $d_{\text{goal}}^{\text{future}}$ refers to the approximate distance to the goal after tracking the twist command for a small amount of time~$\delta t=0.05$~\unit{\second}. This is calculated based on the linearized integral of the robot displacement in the base frame:
\begin{equation}
\begin{split}
     \delta x = v_x^c \delta t - 0.5 v_y^c \omega_z^c \delta t^2, \\
     \delta y = v_y^c \delta t + 0.5 v_x^c \omega_z^c \delta t^2.
\end{split}
\end{equation}
In our practice, gradient descent with a Lagrangian multiplier on the constraint can solve \Cref{eq:ra_rec_solve} within 5 steps when initialized with the current twist, thereby enabling real-time deployment. A visualization of the twist optimization process is given in \Cref{fig:ABS-casestudy} where the searched twist consistently satisfies the safety constraint (i.e., $\hat{V} < V_\text{threshold}$).

\section{Learning Recovery Policy}
\label{sec:recoverypolicy}

The recovery policy is intended to make the robot track a given twist command as fast as possible so that it can function as a backup shielding policy, as mentioned in \Cref{sec:reachavoidvalues}.

\subsection{Observation Space and Action Space}
The observation space of the recovery policy differs from the agile policy in that it tracks twist commands and it does not need exteroception. The recovery policy's observation $o^{\text{Rec}}$ consists of: the foot contacts $c_f$, the base angular velocities $\omega$, the projected gravity in the base frame $g$, the twist commands $tw^c$ (only non-zero variables), the joint positions $q$, the joint velocities $\dot{q}$, and the actions $a$ of the previous frame.

The action space of the recovery policy is exactly the same as that of the agile policy: the 12-d joint targets. We also use an MLP as the policy network.

\subsection{Rewards}
Similar to the agile policy, the reward functions for the recovery policy also consist of three parts: the penalty rewards, the task rewards, and the regularization rewards. The regularization rewards and the penalty rewards remain the same, except that we allow knee contacts with the ground for maximum deceleration (e.g., \Cref{fig:firstpage} (a)).

The task rewards are for twist tracking: 
\begin{equation}
    r_{\text{task}}= 10\cdot r_{\text{linvel}} - 0.5\cdot r_{\text{angvel}} + 5\cdot r_{\text{alive}} -0.1\cdot r_{\text{posture}},
\end{equation}
\emph{i.e.}, a term for tracking $v_x^c$ and $v_y^c$, a term for tracking $\omega_z^c$, a term for staying alive, and a term for maintaining a posture to 
seamlessly switch back to the agile policy. 

To be specific, we have
\begin{equation}
    r_{\text{linvel}} = \exp{\left[-\frac{(v_x-v_x^c)^2+(v_y-v_y^c)^2}{\sigma_{\text{linvel}}^2}\right]},
\end{equation}
where we set $\sigma_{\text{linvel}}=0.5$~\unit{\meter/\second}. For the angular velocity,
\begin{equation}
    r_{\text{angvel}} = \lVert\omega_z - \omega_z^c\rVert_2^2,
\end{equation}
which provides a softer landscape near the command than $r_{\text{linvel}}$. The alive term is simply
\begin{equation}
    r_{\text{alive}} = 1 \cdot \mathds{1}(\text{alive}).
\end{equation}
The posture term is 
\begin{equation}
    r_{\text{posture}} = \lVert q-\bar{q}_{\text{rec}} \rVert_1 ,
\end{equation}
where $\bar{q}_{\text{rec}}$ is a nominal standing pose with low height allowing the robot to switch back to the agile policy seamlessly.

\subsection{Training in Simulation}
The simulation settings for training the recovery policy are similar to those for the agile policy. The differences lie in:
\subsubsection{Domain Randomization}
The observation noises and the dynamic randomization do not change. The episode length is changed to 2~\unit{\second}, and there are randomized initial roll and pitch angles subject to $\mathcal{U}(-\pi/6,\pi/6)$ rad. The randomization ranges are also changed for initial $v_x\sim\mathcal{U}(-0.5,5.5)$~m/s and initial $\omega\sim\mathcal{U}(-1.0,1.0)$~rad/s. These changes better accommodate the states that can trigger the recovery policy during the agile running. The ranges of sampling commands are $v_x^c\sim\mathcal{U}(-1.5,1.5)$~m/s, $v_y^c\sim\mathcal{U}(-0.3,0.3)$~m/s, and $\omega_z^c~\sim\mathcal{U}(-3.0,3.0)$~rad/s.
\subsubsection{Curriculum} The curriculum still exists for terrains and domain randomization. However, the assignment is changed: the robot gets promoted if the velocity tracking error is smaller than $0.7\sigma_{\text{linvel}}$, and gets demoted if it falls over.

\section{Perception}
\label{sec:perception}

As mentioned in \Cref{sec:agilepolicy} and \Cref{sec:reachavoidvalues}, both the agile policy and the RA value network use the exteroceptive 11-d ray distances as part of the observations, with access to their ground truth values during training. These rays are horizontally cast from the robot base, with directions evenly spaced in $[-\frac{\pi}{4},\frac{\pi}{4}]$.

However, such ray distances are not directly available during deployment, and we need to train a ray-prediction network to predict them from depth images, as mentioned in \Cref{sec:systemstructure}.
Such a design leads to the following benefits:
\begin{enumerate}
    \item We only need to tune the ray-prediction network to handle high-dimensional image noises by data augmentation.
    \item The representation is highly interpretable, allowing humans to supervise.
    \item The agile policy and the RA value network are easier to train with low-dimensional inputs.
    \item Compared to costly image rendering in simulation, the ray distances are easy to compute and save training time.
\end{enumerate}
Besides, although ray distances are similar to sparse LiDAR readings, we use cameras instead of LiDARs because a lightweight low-cost camera can easily reach a high FPS, which is important in high-speed collision avoidance.

We present details for training the ray-prediction network in this section.

\subsection{Data Collection}
\begin{figure}[t]
    \centering
    \includegraphics[width=0.65\columnwidth]{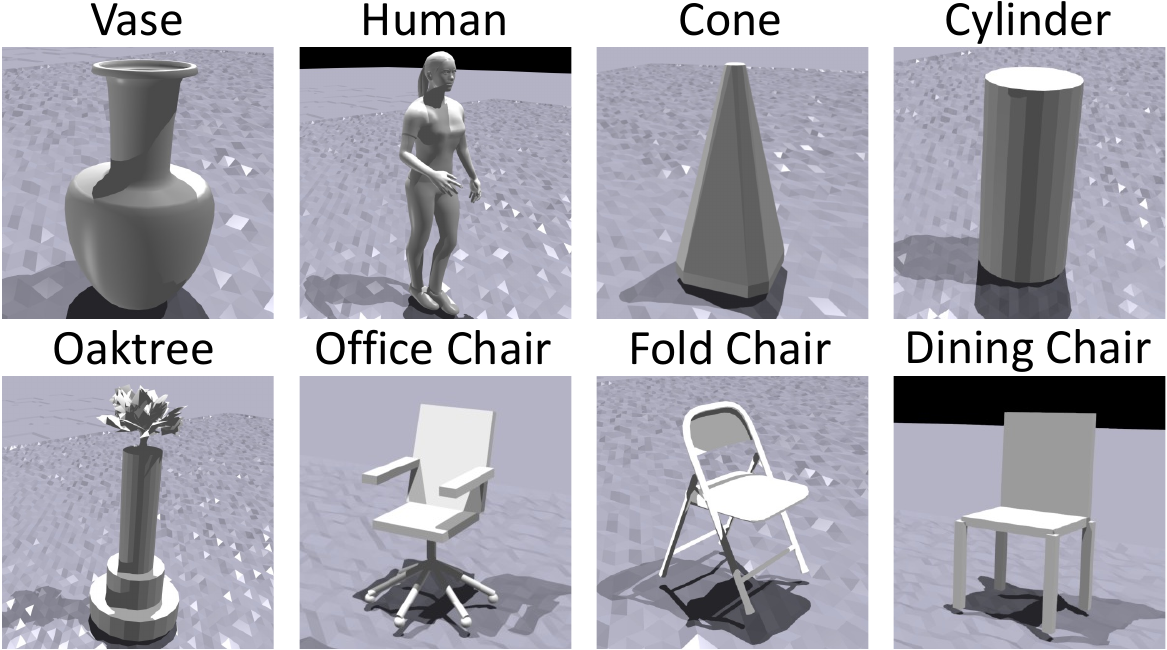}
    \caption{Various obstacles used for ray-prediction data collection.}
    \label{FIG:8obstalces}
\end{figure}
 \begin{figure}[t]
    \centering
    \includegraphics[width=1.0\columnwidth]{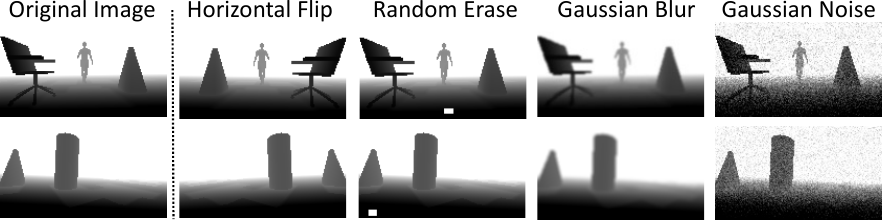}
    \caption{Illustration of four kinds of image augmentation used for depth-based ray-prediction training.}
    \label{FIG:perception_augmentation}
    \vspace{-2mm}
\end{figure}
To train our ray-prediction network, we collect a dataset of pairs of depth images and ray distances (as shown in \Cref{fig:ABS} (a)) by running the agile policy in simulation. The ray-prediction network can then be trained in a supervised way. To facilitate generalization, as shown in \Cref{FIG:8obstalces}, we replace the cylinders with objects of different shapes during data collection.

\subsection{Data Augmentation for Sim-to-Real Transfer}

The real-world depth images collected from cameras are far more noisy than the rendered depth images in simulation~\cite{hoeller2021learning}.
To make the ray-prediction network adapt better to real-world depth images, we apply four data augmentation techniques during training, as shown in \Cref{FIG:perception_augmentation}: 1) horizontal flip; 2) random erase; 3) Gaussian blur; 4) Gaussian noise. 
For deployment, we apply hole filling~\cite{StereolabsFillMode} to further reduce the gap of depth images between the simulation and the real world.

\begin{table*}[t]
\centering
\caption{Benchmarked Comparison in Simulation}
\label{TAB:simcomparison}
\begin{tabular}{cccccc}
\hline
 & Success Rate (\%) & Collision Rate (\%) & Timeout Rate (\%) & $\bar{v}_{\text{peak}}$ of Success (m/s) & $\bar{v}$ of Success (m/s) \\
 \hline
ABS-a & 78.9$\pm$1.4 & 4.4$\pm$0.5 & 16.7$\pm$1.9 & 3.74$\pm$0.02 & 2.15$\pm$0.04\\
ABS-n & 79.1$\pm$4.4 & 5.7$\pm$2.9 & 15.2$\pm$2.1 & 3.48$\pm$0.06 & 2.08$\pm$0.01 \\
ABS-c & \textbf{85.8$\pm$5.6} & \textbf{2.9$\pm$0.7} & 11.3$\pm$5.1 & 2.98$\pm$0.12 & 1.87$\pm$0.03\\ \hline
$\pi^{\text{Agile}}$-a & 73.3$\pm$4.3 & 26.1$\pm$4.4 & \textbf{0.6$\pm$0.1} & \textbf{3.83$\pm$0.03} & \textbf{2.55$\pm$0.03} \\
$\pi^{\text{Agile}}$-n & 77.3$\pm$4.2 & 21.7$\pm$3.9 & 1.0$\pm$0.4 & 3.55$\pm$0.04 & 2.39$\pm$0.04 \\
$\pi^{\text{Agile}}$-c & \textbf{83.2$\pm$1.7} & 15.5$\pm$2.0 & 1.3$\pm$0.6 & 3.04$\pm$0.13 & 2.04$\pm$0.08 \\ \hline
LAG-a & \textbf{82.5$\pm$6.0} & 10.9$\pm$2.6 & 6.6$\pm$4.5 & 2.70$\pm$0.13 & 1.69$\pm$0.09\\
LAG-n & 77.4$\pm$11.5 & 9.1$\pm$1.8 & 13.5$\pm$13.0 & 2.45$\pm$0.07 & 1.41$\pm$0.03 \\
LAG-c & 49.1$\pm$8.4 & 7.4$\pm$2.7 & 43.5$\pm$11.1 & 2.45$\pm$0.10 & 1.12$\pm$0.08 \\
\hline
\multicolumn{6}{l}{*Bold values: the mean falls within the range of top1's mean $\pm$ top1's std.}
\end{tabular}%
\vspace{-2mm}
\end{table*}

\subsection{Other Implementation Details}

To make the network focus more on close obstacles, we take the logarithm of depth values as the NN inputs, and the logarithm of ray distances as the outputs, with the mean squared error as the loss function.

We finetune ResNet-18~\cite{he2016deep} with pretrained weights to train the model. The images are downsampled to $[160,90]$ resolution both in simulation and during deployment.

\section{Experiments}

\subsection{Baselines}
For experimental results, we consider three settings:
\begin{enumerate}
    \item Our \method system, with both the agile policy and the recovery policy;
    \item Our agile policy $\pi^{\text{Agile}}$ only;
    \item ``LAG": we use PPO-Lagrangian~\cite{OpenAI2019SafeRL} to train end-to-end safe RL policies with the agile policy's formulation.
\end{enumerate}

By comparing (2) and (3), we can see how agility and safety trade off without external modules, forming a boundary of agility and safety. With the help of RA values and the recovery policy, we expect (1) to break this boundary with a high safety gain: it should be as agile as (2) in safe cases, and shield the robot in risky cases.

\textit{Note that the three settings here are all based on what we propose. A detailed comparison between our agile policy and the previous state-of-the-art (SOTA) agile running policy~\cite{margolis2022rapid} is made in \Cref{subsubsec:goal_vs_vel}.}

\subsection{Simulation Experiments}
\label{sec:simulationexperiments}

\subsubsection{Quantitative results}
We test the policies trained with different settings in simulation. To better show the agility-safety boundary, we introduce 3 variants for each setting: an aggressive one (``-a") doubling the agile reward term $r_{\text{agile}}$, a nominal one (``-n"), and a conservative one (``-c") halving the $r_{\text{agile}}$. Regarding the obstacles, we distribute eight obstacles within a 5.5~\unit{\metre}$\times$ 4~\unit{\meter} rectangle (during training it was 11~\unit{\metre}$\times$ 5~\unit{\meter}), so the test cases are in distribution but much harder than most cases during training. 

\begin{figure}[t]
\centering
\includegraphics[width=0.9\linewidth, trim={0cm 0 0 0},clip]{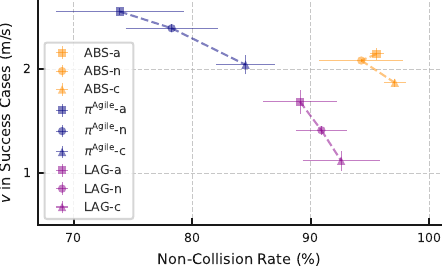}
\vspace*{-3mm}
\caption{Illustration of agility-safety trade-off in benchmarked comparison. Agility is quantified by the average speed achieved in success cases while safety is represented by the non-collision rate. Points indicate the mean values, and error bars indicate the std values.}
\label{FIG:simcomparison}
\vspace{-2mm}
\end{figure}

\begin{figure}[t]
    \centering
    \includegraphics[width=\linewidth]{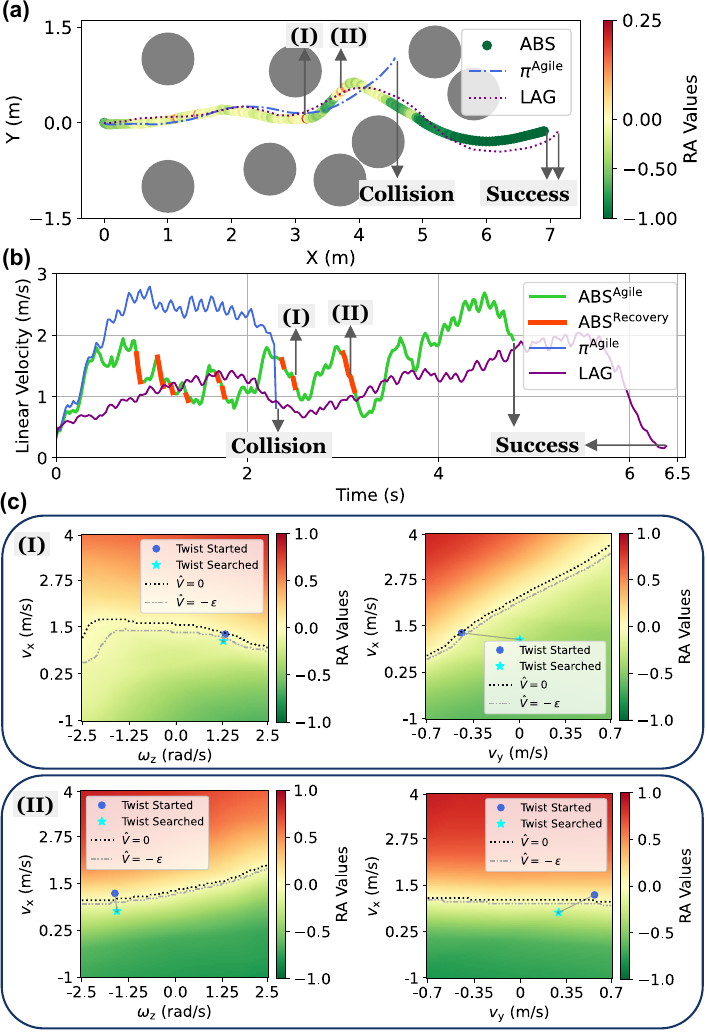}
    \caption{An example case in simulation where $\pi^{\text{Agile}}$ fails to reach the goal. 
    a) Trajectories of \method and other baselines, with RA values visualized for \method. 
    b) The velocity-time curves showing that \method is much faster than the LAG baseline.
    c) Illustrations of the RA value landscape when the recovery policy is triggered at (I) and (II), projected in the $v_x-\omega_z$ plane and the $v_x-v_y$ plane.    
    We show the initial twist before search (\emph{i.e.}, the current twist of the robot base) and the searched commands based on \Cref{eq:ra_rec_solve}.} 
    \label{fig:ABS-casestudy}
    \vspace{-2mm}
\end{figure}

The results are reported in Table~\ref{TAB:simcomparison} and \Cref{FIG:simcomparison}. There are three possible outcomes for an episode: success, collision, or timeout. Trajectories that do not trigger success or collision criteria within the episode length are labelled as ``timeout''.  We report the success rate, the collision rate, the timeout rate, the average peak velocity for success cases, and the average speed for success cases as the metrics. For each setting, the mean and std values are calculated over 3 policies trained with different seeds, and the metrics are obtained via testing for 10k random episodes.

The results indicate that, no matter how the reward weights are tuned or whether the RL algorithm is constrained by safe exploration, the agility and the safety trade off within a boundary. Yet, with our RA values and the recovery policy as a safeguard, we can break this boundary and get a substantial improvement in safety at the cost of only a minor decrease in agility.

\textit{Note that the variants are only introduced to show the boundary here. In the following parts, we will only use the nominal ones.}

\subsubsection{Example Case}

We present an example case of \method and other baselines in simulation, where the robot starting from $(0,0)$ needs to run through $8$ obstacles to reach the goal $(7,0)$, as shown in \Cref{fig:ABS-casestudy}. The robot needs to first go through an open space, followed by two tight spaces, and then another open space. In this case, the $\pi^{\text{Agile}}$ baseline runs fast but \Revise{collides} near the second tight space. The LAG baseline runs much slower than \method. In contrast, our proposed \method runs fast in the open spaces, and slows down in the tight spaces for safety thanks to the shielding of RA values and the recovery policy. \Cref{fig:ABS-casestudy} (c) demonstrates the RA value landscape with respect to twist commands when the recovery policy is activated, where the searched twist consistently satisfies the safety constraint (i.e., $\hat{V} < V_\text{threshold}$).

\subsection{Real-World Experiments}
\label{sec:realworldexperiments}
\subsubsection{Hardware setup}
We use the Unitree Go1 for our experiments. 
The robot is equipped a Jetson Orin NX for onboard computation and a ZED Mini Stereo Camera for depth and odometry sensing.  
We employ the ZED odometry module to online update the relative goal commands for the agile policy, with its difference as the velocity estimation.
We use Unitree's built-in PD controller, with $K_p = 30$ and $K_d = 0.65$.

\subsubsection{Results}

\begin{figure}[t]
    \centering
    \includegraphics[width=1.0\columnwidth]{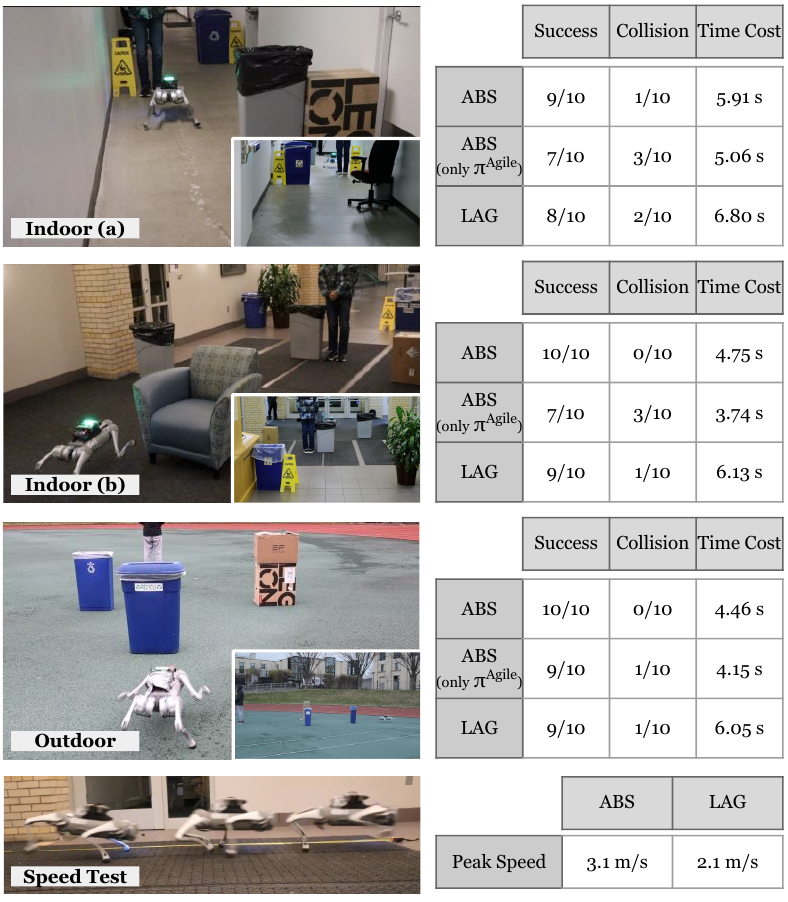}
    \caption{We evaluate \method and other baselines in the real world with two indoor testbeds, one outdoor testbed, and repetitive speed tests. Indoor (a) is a dim and narrow corridor, Indoor (b) is a hall with furnitures, and Outdoor is an open space on the playground with few obstacles. \method achieves the best safety across three testbeds, with faster speeds compared to the LAG baseline.}
    \label{FIG:real-quantitive}
    \vspace{-2mm}
\end{figure}

\begin{figure}[t]
    \centering
    \includegraphics[width=1.0\linewidth]{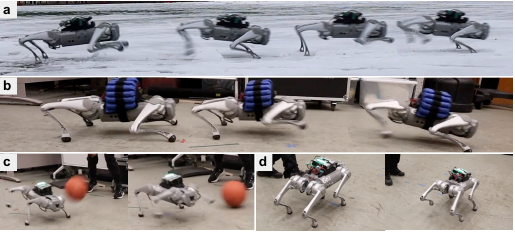}
    \caption{Robustness Tests of our \method system, a) in snowy terrain,s b) bearing a $12$-kg payload, c) against a ball hit when running, and d) withstand a kick when standing at the goal.}
    \label{fig:robustness_test}
    \vspace{-4mm}
\end{figure}

Across two indoor and one outdoor testbeds, \method demonstrates superior overall performance as shown in \Cref{FIG:real-quantitive}, achieving the highest success rates and the lowest collision rates. Specifically, \method consistently scores either 9 or 10 out of 10 in success rates across all environments, with minimal collisions, indicating robustness and reliability in the real world. 

Without the safety shield, the agile policy $\pi_{\text{Agile}}$ achieves the fastest running speed at the cost of more collisions. LAG outperforms $\pi_{\text{Agile}}$ in safety but has slower speeds, and falls short in both safety and agility compared to \method. \method achieves high speed with high safety, and generalizes to dynamic obstacles, as shown in \Cref{fig:firstpage}.

\subsubsection{Robustness}
Our \method system can work on the slippery icy snow, bear a $12$-kg payload (equal to its own weight), and withstand perturbations, as shown in \Cref{fig:robustness_test}. These tests demonstrate the robustness of our system.

\section{Extensive Studies and Analyses}

\subsection{Maximizing Agility}
\label{sec:agilelocomotionanalysis}

\subsubsection{Goal-Reaching v.s. Velocity-Tracking}
\label{subsubsec:goal_vs_vel}

Velocity-tracking is the most commonly used formulation of locomotion controllers~\cite{rudin2022learning,margolis2022rapid,miki2022learning,lee2020learning}, and is also adopted for our recovery policy. However, for the agile policy, we claim that goal-reaching is a better choice because it does not decouple locomotion and navigation for collision avoidance and can fully unleash the agility that is learned. Moreover, we empirically find that the goal-reaching formulation benefits sim-to-real transfer as it finds a better gait pattern for high-speed running.

With the SOTA agile velocity-tracking policy in~\cite{margolis2022rapid} as a baseline (referred to as ``rapid''), we make detailed comparisons in Table~\ref{TAB:posvsvel}. For fair comparison, we train ``rapid'' with the combination of our regularization rewards and the task rewards in~\cite{margolis2022rapid}, use the same action space for two policies, and remove the temporal information and system identification in~\cite{margolis2022rapid}.

\begin{table}[t]
    \centering
    \caption{Goal-Reaching Policy v.s. Velocity-Tracking Policy}
    \label{TAB:posvsvel}
    \resizebox{1.0\columnwidth}{!}{
    \begin{tabular}{ccc}
    \hline
      Term & Our $\pi^{\text{Agile}}$ & Rapid~\cite{margolis2022rapid}\\
      \hline
      Gait patterm & \textbf{gallop} & near trot \\
      Max \#. uncontrollable DoFs & $\mathbf{1}$ & $3$ \\
      \hline
      Peak vel. in simulation & $4.0$~m/s & \textbf{$\mathbf{4.1}$~m/s} \\
      Peak torque in simulation & \textbf{$\mathbf{23.5}$~Nm} & 35.5~Nm \\
      Peak joint vel. in simulation &\textbf{$\mathbf{22.0}$~rad/s} & 30.0~rad/s \\
      Peak vel. in real world & \textbf{$\mathbf{3.1}$~m/s} & $2.5$~m/s \\
      \hline
      Collision avoidance & \textbf{as trained }& need high-level commands \\
      Fully unleashed agility &\textbf{ as trained}  & non-trivial for high level \\
      Changing vel. for steering & \textbf{in distribution} & out of distribution \\
      Curriculum learning & \textbf{straightforward} & carefully designed \\
         \hline
    \end{tabular}}
    \vspace{-2mm}
\end{table}

\subsubsection{Effects of illusion and ERFI-50 randomization}
\begin{figure}[t]
    \centering
    \includegraphics[width=1.0\linewidth]{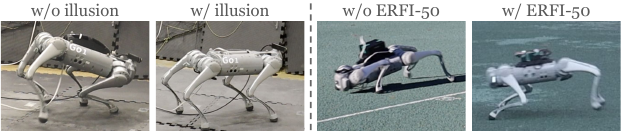}
    \caption{Effects of illusion and ERFI-50 randomization. The robot will tremble near a wall without illusion randomization and will hit the ground during running without ERFI-50 randomization.}
    \label{fig:illusion-erfi}
    \vspace{-4mm}
\end{figure}

Two key components we add in domain randomization to help sim-to-real transfer is the illusion and the ERFI-50. As shown in \Cref{fig:illusion-erfi}, without the illusion, the robot will sometimes tremble near a wall which it has never seen in simulation. Without ERFI-50, the robot will hit the ground with its head during running due to the sim-to-real gap in motor dynamics.

\Revise{
\subsection{Extensive Studies on RA Values}

\subsubsection{Selecting safety threshold} For safety shielding, we choose $V_{\text{threshold}}=-0.05$. Theoretically, this threshold captures the conservativeness of the switching strategy and the discrepancy between the true reach-avoid value and its learned approximation. 
Our framework, however, demonstrates robustness to the selection of $V_{\text{threshold}}$, as evidenced by \Cref{TAB:vthres} where scanning $V_{\text{threshold}}$ from $-0.001$ to $-0.1$ brings no significant change in the overall performance. For a big value ($0.1$ is considered big given that $\hat{V}$ is bounded between $-1$ and $1$), the collision rate slightly decreases as expected whereas the success rate also slightly decreases.

}
\begin{table}[t]
\centering
\caption{\Revise{Effects of Different $V_{\text{threshold}}$ on ABS}}
\label{TAB:lipablate}
\resizebox{\linewidth}{!}{
\begin{tabular}{ccccc}
\hline
 $V_{\text{threshold}}$ & -0.001 & -0.01 & -0.05  & -0.1\\
 \hline
Success Rate (\%) & 78.0$\pm$2.1 & 78.1$\pm$3.4 & 79.1$\pm$4.4 & 75.8$\pm$ 2.0 \\
Collision Rate (\%)& 5.0$\pm$0.6 & 5.8$\pm$1.9 & 5.7$\pm$2.9& 4.3$\pm$0.6 \\
$\bar{v}_{\text{peak}}$ of Success (m/s) & 3.42$\pm$0.06 & 3.46$\pm$0.08 & 3.48$\pm$0.06 & 3.42$\pm$0.05\\
$\bar{v}$ of Success (m/s) & 2.08$\pm$0.02 & 2.08$\pm$0.01 & 2.08$\pm$0.01 & 2.05$\pm$ 0.03\\
\hline
\end{tabular}%
}
\vspace{-2mm}
\end{table}

\Revise{
\subsubsection{Soft Lipschitz continuity for the failure indicator} In \cite{fisac2015reach}, the Lipschitz continuity is used to prove the existence and uniqueness of a solution to the value function. In our paper, we soften the discrete collision indicator to approach the Lipschitz continuity, and ablate its effects here. As shown in \Cref{TAB:lipablate}, this technique significantly enhances the safety of our system while slightly increasing the conservativeness, corroborating the observations made in \Cref{FIG:RA-POS_visualization}.}

\begin{table}[t]
\centering
\caption{\Revise{Effects of Softened Failure Indicator on ABS}}
\label{TAB:vthres}
\resizebox{\linewidth}{!}{
\begin{tabular}{cccc}
\hline
 & ABS w/ softened $\zeta$ & ABS w/o softened $\zeta$ & $\pi^{\text{Agile}}$ \\
 \hline
Success Rate (\%) & 79.1$\pm$4.4 & \textbf{81.7$\pm$1.3} & 77.3$\pm$4.2\\
Collision Rate (\%)& \textbf{5.7$\pm$2.9} & 14.7$\pm$1.5 & 21.7$\pm$3.9\\
$\bar{v}_{\text{peak}}$ of Success (m/s) & 3.48$\pm$0.06 & 3.45$\pm$0.06 & \textbf{3.55$\pm$0.04}\\
$\bar{v}$ of Success (m/s) & 2.08$\pm$0.01 & 2.27$\pm$0.03 & \textbf{2.39$\pm$0.04}\\
\hline
\end{tabular}%
}
\vspace{-2mm}
\end{table}

\Revise{
\subsubsection{Can RA values shield LAG baseline} Given that our framework is general in shielding the goal-reaching policy with RA values and the recovery policy, it can also shield the LAG baseline, which makes a variant of \method with a cost-critic. We present the results of shielded LAG in \Cref{TAB:lagra}, showcasing improved safety over LAG despite slightly reduced agility. The LAG+RA setting can still reach an average speed of $>1$~m/s, demonstrating the power of our \method framework compared to existing works.
}

\begin{table}[t]
\centering
\caption{\Revise{Effects of RA Shielding on LAG}}
\label{TAB:lagra}
\begin{tabular}{cccc}
\hline
 & ABS & LAG & LAG+RA \\
 \hline

Success Rate (\%) & \textbf{79.1$\pm$4.4} & 77.4$\pm$11.5 & 70.5$\pm$7.7\\
Collision Rate (\%)& 5.7$\pm$2.9& 9.1$\pm$1.8 & \textbf{2.8$\pm$1.2} \\
$\bar{v}_{\text{peak}}$ of Success (m/s) & \textbf{3.48$\pm$0.06} & 2.45$\pm$0.07 & 2.40$\pm$0.11\\
$\bar{v}$ of Success (m/s) & \textbf{2.08$\pm$0.01} & 1.41$\pm$0.03 & 1.22$\pm$0.08 \\
\hline
\end{tabular}%
\vspace{-2mm}
\end{table}

\begin{table}[t]
\centering
\caption{\Revise{Performance Metrics for Different Network Architectures and Training Approaches}}
\label{TAB:Perception}
\resizebox{\linewidth}{!}{%
\begin{threeparttable}[flushleft]
\begin{tabular}{lcc}
\toprule
\textbf{Architecture} & \textbf{Test Set MSE} & \textbf{Inference Time (ms)} \\
\midrule
EfficientNet-B0* & $ 3.627 \times 10^{-2}$ & 19 \\
MobileNet-V2* & $3.387 \times 10^{-2}$ & 15 \\
ResNet-34 & $3.081 \times 10^{-2}$ & 14 \\
ResNet-18 & $3.238 \times 10^{-2}$ & 9 \\
ResNet-18 (w/o pretraining) & $3.526 \times 10^{-2}$ & 9 \\
ResNet-18 (w/o augmentation) & $3.393 \times 10^{-2}$ & 9 \\
\bottomrule
\end{tabular}%
\begin{tablenotes}
\item[*] {\scriptsize We use the PyTorch-ONNX pipeline where the implementations of these network architectures may be not fully optimized.}
\end{tablenotes}
\end{threeparttable}
}
\vspace{-1mm}
\end{table}

\begin{figure}[t]
    \centering
    \includegraphics[width=1.0\linewidth,trim=3.5cm 1cm 0cm 1cm,clip]{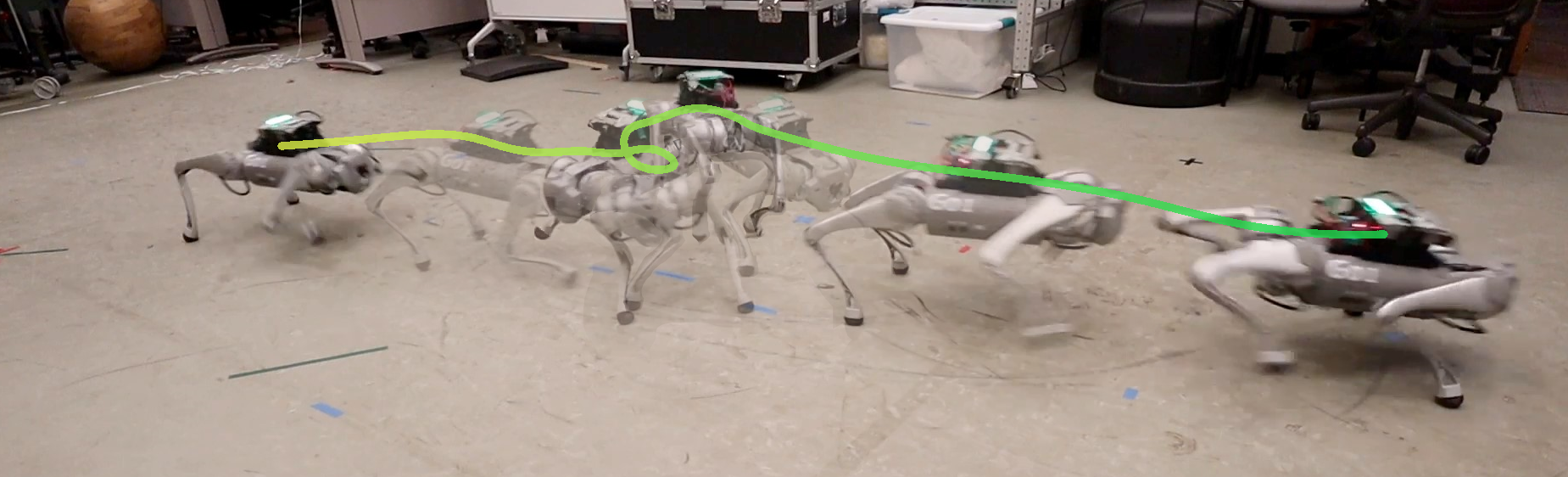}
    \caption{Steering the robot with a command sequence ``Forward"-``Rapid Right Turn"-``Forward". The robot can reach $>3$~m/s when running forward and $>6$ rad/s when turning rapidly.}
    \label{fig:run_rot_run}
    \vspace{-2mm}
\end{figure}

\begin{table}[t]
\centering
\caption{Goal Commands for Instant Steering}
\label{tab:steer2goal}
\resizebox{\columnwidth}{!}{%
\begin{tabular}{lccc}
\hline
Steering         & Goal $x$ (m) & Goal $y$ (m) & Goal Heading (rad) \\ \hline
Forward          &   $5$    &     $0$   &     $0$     \\
Stop             &  $0$ & $0$ & $0$ \\
Left Turn        &   $2$  &    $1.5$  &      $\frac{\pi}{2}$   \\
Rapid Left Turn  &   $-2$    &   $ 0 $    &  $3$  \\
Right Turn       &   $2$    &   $-1.5$    &   $-\frac{\pi}{2}$     \\
Rapid Right Turn &   $-2$     &   $ 0 $   &   $-3$     \\ \hline
\end{tabular}%
}
\vspace{-4mm}
\end{table}

\subsection{Enhancing Perception Training}
\label{sec:perceptiontraining}

In refining our ray-prediction network training, we systematically examine several factors: 1) network architecture, 2) pretrained weights, and 3) data augmentation. The comparative results, detailed in \Cref{TAB:Perception}, underscore the significance of both pretrained weights and data augmentation in enhancing the accuracy of the network.

Regarding the network size, larger networks improve the prediction accuracy at the cost of inference time. Note that the reported inference time in \Cref{TAB:Perception} was measured on Jetson Orin NX exclusively for perception inference. In practical deployment where computational resources are shared among various tasks, the actual update frequency can be considerably lower. 
For real-time high-speed locomotion, we opt for ResNet-18, balancing accuracy and responsiveness in dynamic environments.


\subsection{Instant Steering via Commands}
\label{subsec:insSteer}

As mentioned in \Cref{sec:agilepolicy}, we can change our goal commands even in the run time. Thanks to our single-frame observations and randomization settings, we can easily overwrite goal commands to achieve instant agile steering, as presented in \Cref{tab:steer2goal}. This enables direct human involvement similar to the velocity-tracking formulation, and \Cref{fig:run_rot_run} showcases such operations in the real world.

\subsection{Failure Cases and Limitations}
\label{sec:failurecases}
First, when the obstacles are too dense and form a local minimum, our policy can easily fail, which is also reflected by the high timeout rates in the results. This is common in local navigation planners~\cite{zhang2024resilient, mattamala2022efficient} though, and a potential solution can be to add memory~\cite{wijmans2023emergence} or introduce a global hint~\cite{truong2023i2o}.

Second, our generalization to dynamic environments are due to the shielding of RA values and the recovery policy. The RA values are learned with static obstacles, and can only generalize to quasi-static environments.
If a dynamic object moves faster than the velocity limit of the recovery policy, the collision may happen. A potential solution is to predict the motions of the obstacles in the future~\cite{large2004avoiding, si2019agen}.

Third, we limit the robot behaviors to only 2D locomotion and constrain the motions to have no flying phase. For 3D terrains such as stairs and gaps, the problem can be far more challenging because the locomotion skills and the collision avoidance are coupled.

Fourth, implicit system identification techniques~\cite{lee2020learning, miki2022learning, margolis2022rapid, kumar2021rma} can leverage temporal information to represent real-world dynamics and facilitate sim-to-real transfer, but it is non-trivial to incorporate them into our system. 
This requires a latent embedding of the temporal information, which is hard to deal with for the RA module. The policy switch can also make the embedding out of distribution for the policies.

Fifth, the vision system needs further improvement. In the Indoor (a) testbed, the only collision of \method is due to the ``undetected'' objects by the ray-prediction network as the corridor is quite dim. Beside the network, the system can also be completed by adding more cameras around the body. In this way, the robot may also dodge the obstacles come from behind or the side. Event cameras may also help in highly dynamic scenarios, e.g., when dodging a high-speed ball~\cite{falanga2020dynamic}

\section{Conclusion} 
\label{sec:conclusion}
In this paper, we achieve safe high-speed quadrupedal locomotion in cluttered environments. Our framework \method employs a dual-policy setup where the agile policy enables the robot to run fast, and the recovery policy safeguards the robot. The learned reach-avoid values govern the policy switch and guide the recovery policy. A ray-prediction network provides exteroception representation for the agile policy and the RA value network. Some key takeaways are:
\Revise{\begin{enumerate}
    \item \textbf{Advanced agility:} Decoupled locomotion and navigation constrains the agility of collision-free locomotion in existing works. In contrast, we train an end-to-end agile policy with the local navigation formulation to fully unleash the agility while avoiding obstacles.
    \item \textbf{Safeguarded agility:} The agile policy does not guarantee safety, while using the Lagrangian-based RL or adjusting reward weights only trades off agility and safety. In this paper, we use external shielding modules to help break the trade-off boundary.
    \item \textbf{Model-free safety:} Model-based methods explicitly enforce safety through constraints but struggle with complex dynamics and high-dimensional states. In this paper, we learn RA values in a model-free way and condition them on the policy to simplify learning and enable offline learning from policy rollouts.
    \item \textbf{Guided recovery:} Designing a recovery policy capable of protecting the robot from potential collisions across various obstacle distributions and states is non-trivial. In this paper, we guide the recovery policy with RA values and their gradients to ensure safety.
    \item \textbf{Perception to motion:} Visual inputs are high-dimensional and noisy, while efficient learning of robust controllers is desired. To this end, we use a low-dimensional exteroception representation to facilitate policy learning and generalization, and train a ray-prediction network to map raw depth images to the representation.
\end{enumerate}
}
\section*{Acknowledgments}
We appreciate Wennie Tabib for supporting hardware experiments, and thank Yuxiang Yang, Yiyu Chen, Yikai Wang and Xialin He for their advice on hardware debugging, and Arthur Allshire for help on software debugging. Special thanks to Andrea Bajcsy and Ziqiao Ma for their assistance in graphics design. This work was in part supported by NSF under grant No. 2144489.


\bibliographystyle{plainnat}
\bibliography{references}

\clearpage
\Revise{
\section*{Appendix}
\subsection{Training Costs}
On an NVIDIA RTX 4090, each iteration of the agile policy learning takes 1.5~sec. The $\pi^{\text{Agile}}$ policy needs $\sim800$ iterations ($\sim 20$~min) to converge, and the LAG policy needs $\sim 2500$ iterations ($\sim 1$~h) to converge. Yet, to fully unleash their performance and make a fair comparison, all of the policies are tested with the checkpoints of $10000$ iterations ($\sim 4$~h). 
The recovery policy is far more efficient and converges within $500$ iterations, becoming near-optimal within 10~min. The GPU memory usage during training is below 5GB.

We train the RA value network in parallel with the agile policy rollout to save time, and the procedure takes $\sim 40$~min. 

The ray-prediction network can use collected data from different trained policies as the prediction in ideal cases is not conditioned on the policy. The network training time can vary for different settings. As a reference, we collected 250k labelled images for training.

\subsection{Obtaining Ray Distances in Simulation}
Ideally, we should use ray tracing tools such as Embree~\cite{embree} and Warp~\cite{nvidia_warp} to obtain accurate ray distances. However, these tools are time-consuming and GPU-intensive for our separated parallel environments with multiple objects. To enable efficient training, we implement an analytical ray tracing calculation, which, while efficient, is limited to simple geometries such as cylinders. Therefore, we are unable to use diverse objects during training, and have to choose the objects with meshes that can be approximated during ray-prediction data collection.

Despite this limitation, our design of the low-dimensional exteroception representation, together with domain randomization and data augmentation, enables the system to generalize to a variety of real-world objects, as demonstrated in the Figure~\ref{fig:firstpage}.
}
\Revise{
\subsection{RL Hyperparameters} }
\begin{table}[h]
\centering
\caption{\Revise{RL Hyperparameters}}
\label{tab:hpparam}
\begin{tabular}{cc}
\hline
Hyperparameter & Value \\ \hline
\textbf{Agile Policy} & \\
\#. steps per iteration & 48 \\
Entropy coefficient & 0.003 \\
Actor NN & MLP with hidden units [512, 256, 128] \\
Critic NN  & MLP with hidden units [512, 256, 128] \\
Others & Same as \cite{rudin2022learning} \\ \hline
\textbf{Agile Policy - LAG} & \\
Cost-critic NN & MLP with hidden units [512, 256, 128] \\
Cost-critic loss coefficient & 1.0 \\
Cost gamma & 0.99 \\
Cost lambda & 0.97 \\ 
Cost limit & 0.0 \\ 
Multiplier learning rate & 0.001 \\ \hline
\textbf{Recovery Policy} & \\
\#. steps per iteration & 24 \\
Entropy coefficient & 0.003 \\
Actor NN & MLP with hidden units [512, 256, 128] \\
Critic NN  & MLP with hidden units [512, 256, 128] \\
Others & Same as \cite{rudin2022learning} \\ 
\hline
\end{tabular}%
\end{table}

\Revise{
\subsection{Explanations for Regularization Terms}
We explain why we add so many regularization terms in Table~\ref{tab:regexplain}, which follows \cite{rudin2022learning, rudin2022advanced, margolis2022rapid, zhang2023learning, hoeller2023anymal} and has some adaptations based on our tuning experiences.
}
\begin{table}[h]
\centering
\caption{\Revise{Explaining Regularization Rewards}}
\label{tab:regexplain}
\begin{tabular}{cp{3.9cm}}
\hline
Term & Reason \\ \hline
 $v_z^2$ & To reduce vertical oscillation \\
 $\omega_x^2+\omega_y^2$ & To reduce rotational oscillation \\
 $(g_x^2+g_y^2)$ & To reduce tilting \\ 
 $ \lVert \tau \rVert_2^2$ & To reduce mechanical stress and power consumption of motions \\
 $\sum\nolimits_{i=1}^{12} {\rm{ReLU}}\left(\left|\tau_{i}\right|-0.85\cdot\tau_{i,\lim }\right)$ & To avoid large torques that can lead to sim-to-real gaps \\
 $\lVert \dot{q} \rVert_2^2 $ & To reduce aggressive motions that may damage the hardware \\
 $\sum\nolimits_{i=1}^{12} {\rm{ReLU}} \left(\left|\dot{q}_{i}\right|-0.9\cdot\dot{q}_{i,\lim }\right) $ & To avoid large joint velocities that can lead to inaccurate simulation \\
 $\sum\nolimits_{i=1}^{12} {\rm{ReLU}} \left(\left|{q}_{i}\right|-0.95\cdot{q}_{i,\lim }\right)$ & To avoid joint positions near the limit that can lead to inaccurate simulation  \\
 $\lVert \ddot{q}\rVert_2^2$ & To reduce jerky motions \\
 $\lVert \dot{a}\rVert_2^2$ & To encourage smooth actions \\ 
 $\mathds{1}(\text{fly})$ & To constrain \#. uncontrollable DoFs and improve maneuverability \\
 
 \hline
\end{tabular}%
\end{table}

\end{document}